\pdfoutput=1

\documentclass[11pt]{article}

\usepackage[]{acl}

\usepackage{times}
\usepackage{latexsym}
\usepackage{soul}
\usepackage{graphicx}
\usepackage{booktabs}
\usepackage{multirow}
\usepackage{enumitem}

\usepackage[T1]{fontenc}

\usepackage[utf8]{inputenc}

\usepackage{microtype}

\usepackage{color}
\usepackage{booktabs}
\usepackage{algorithm}
\usepackage{algorithmic}
\usepackage{amsmath}

\newcommand{\githublink}{\url{https://github.com/salesforce/dialog-flow-extraction}}

\title{Structure Extraction in Task-Oriented Dialogues with Slot Clustering}

\author{Liang Qiu$^\dag$, Chien-Sheng Wu$^\ddag$, Wenhao Liu$^\ddag$, Caiming Xiong$^\ddag$ \\
  UCLA Center for Vision, Cognition, Learning, and Autonomy$^\dag$ \\    
  Salesforce AI Research$^\ddag$ \\
  \texttt{liangqiu@ucla.edu, \{wu.jason, wenhao.liu, cxiong\}@salesforce.com} \\}

\begin{document}
\maketitle
\begin{abstract}
Extracting structure information from dialogue data can help us better understand user and system behaviors. In task-oriented dialogues, dialogue structure has often been considered as transition graphs among dialogue states. However, annotating dialogue states manually is expensive and time-consuming. In this paper, we propose a simple yet effective approach for structure extraction in task-oriented dialogues. We first detect and cluster possible slot tokens with a pre-trained model to approximate dialogue ontology for a target domain. Then we track the status of each identified token group and derive a state transition structure. Empirical results show that our approach outperforms unsupervised baseline models by far in dialogue structure extraction. In addition, we show that data augmentation based on extracted structures enriches the surface formats of training data and can achieve a significant performance boost in dialogue response generation\footnote{Data and code are available at \githublink.}. 
\end{abstract}

\section{Introduction}
\begin{figure}[t] 
  \begin{minipage}{0.49\linewidth} 
    \includegraphics[width=1.0\textwidth]{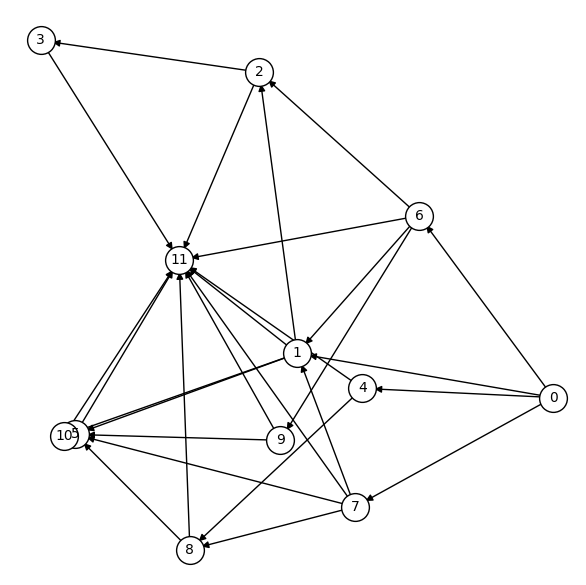}
  \end{minipage}
  \begin{minipage}{0.49\linewidth} 
    \includegraphics[width=1.0\textwidth]{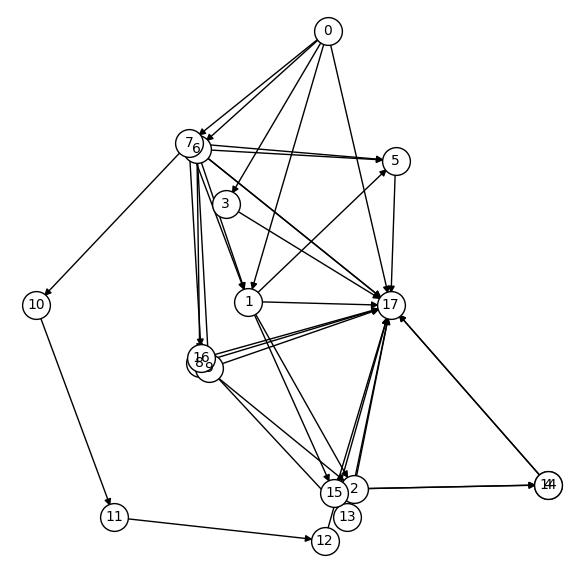}
  \end{minipage}
  \caption{Dialogue structure in the \textit{attraction} domain of the MultiWOZ~\cite{budzianowski-etal-2018-multiwoz}. The structure on the left is from annotated dialogue states, while the right one is extracted by our approach. Structures for other domains are attached in Appendix~\ref{sec:appendix}. } 
  \label{fig:attraction}
\end{figure}

There is a long trend of studying the semantic state transition in dialogue systems. For example, modeling dialogue states in a deep and continuous space has been shown to be beneficial in response generation task~\cite{serban2017hierarchical, wen-etal-2017-network, wu-etal-2020-tod}. While in a modular system~\cite{bocklisch2017rasa} that is more preferred in industry, dialogue states are explicitly defined as the status of a set of slots. The domain-specific slots are often manually designed, and their values are updated through the interaction with users, as shown in Table~\ref{tab:attraction}. Extracting structure information from dialogue data is an important topic for us to analyze user behavior and system performance. It also provides us with a discourse skeleton for data augmentation. Figure~\ref{fig:attraction} shows an example of dialogue structure in the \textit{attraction} domain of the MultiWOZ dataset~\cite{budzianowski-etal-2018-multiwoz}. Each node represents a distinct dialogue state in Table~\ref{tab:attraction}, where the three dialogue turns correspond to node \textcircled{\scriptsize{0}}, \textcircled{\scriptsize{1}} and \textcircled{\scriptsize{2}} respectively. And the edges indicate transitions between pairs of states.
\begin{table}[t]
\resizebox{\linewidth}{!}{
\begin{tabular}{lc}
\toprule
Dialogue                                                                                                                                                                                           & \begin{tabular}[c]{@{}c@{}}Dialogue State\\ Slot Value\end{tabular}                   \\
\midrule
\begin{tabular}[c]{@{}l@{}}{[}usr{]} Can you please help me find a place to go?\\ {[}sys{]} I've found 79 places for you to go. \\Do you have any specific ideas in your mind?\end{tabular} & \begin{tabular}[c]{@{}c@{}}{[}0, 0, 0{]} $\rightarrow$ \textcircled{\scriptsize{0}} \\ {[}`', `', `'{]}\end{tabular}              \\
\midrule
\begin{tabular}[c]{@{}l@{}}{[}usr{]} I'd like a \textbf{sports} place in the \textbf{centre} please.\\ {[}sys{]} There are no results matching your query. \\Can I try a different area or type?\end{tabular} & \begin{tabular}[c]{@{}c@{}}{[}0, 1, 1{]} $\rightarrow$ \textcircled{\scriptsize{1}} \\ {[}`', `sports', `centre'{]}\end{tabular}        \\
\midrule
\begin{tabular}[c]{@{}l@{}}{[}usr{]} Okay, are there any \textbf{cinemas} in the centre?\\ {[}sys{]} We have vue cinema.\end{tabular}                                                                    & \begin{tabular}[c]{@{}c@{}}{[}0, 2, 1{]} $\rightarrow$ \textcircled{\scriptsize{2}} \\ {[}`', `cinemas', `centre'{]}\end{tabular} \\
\bottomrule
\end{tabular}
}
\caption{Example dialogue in the \textit{attraction} domain of the MultiWOZ~\cite{budzianowski-etal-2018-multiwoz}. \textbf{Bold} tokens are detected by our algorithm as potential slots and used to update the dialogue state. The dialogue state vectors record how many times each slot is updated.}
\label{tab:attraction}
\end{table}

However, high-quality dialogue data with complete dialogue state annotation is limited. Existing works put more emphasis on unsupervised learning of dialogue structures. Representative ones include training language models based on Hidden Markov Models (HMMs)~\cite{chotimongkol2008learning} or Variational Recurrent Neural Networks (VRNNs)~\cite{shi-etal-2019-unsupervised, qiu-etal-2020-structured} to reconstruct the original dialogues. The structure built upon the latent states is then evaluated in downstream tasks like dialogue policy learning. Since the latent states are implicitly defined, there is a gap between the learned structure and the canonical dialogue states in task-oriented dialogues, making the structure hard to interpret and analyze. What's more, it remains unclear that how we should choose the number of states during extraction. The state number directly decides the structure granularity, but it is not available in practice. 

To alleviate these problems, we propose a simple yet effective approach for structure extraction in task-oriented dialogues. 
First, we define a task called Slot Boundary Detection (SBD). Utterances from training domains are tagged with the conventional BIO schema but without the slot names. A Transformer-based classifier is trained to detect the boundary of potential slot tokens in the test domain. 
Second, while the state number is usually unknown, it is more reasonable for us to assume the slot number can be estimated by checking just a few chat transcripts. We therefore cluster the detected tokens into groups with the same number of slots. 
Finally, the dialogue state is represented with a vector recording the modification times of every slot. We track the slot values through each dialogue session in the corpus and label utterances with their dialogue states accordingly. The semantic structure is portrayed by computing the transition frequencies among the unique states.

We evaluate our approach against baseline models that directly encode utterances or use rule-based slot detectors, besides the afore-mentioned latent variable model VRNN. Empirical results in the MultiWOZ dataset~\cite{budzianowski-etal-2018-multiwoz} show that the proposed method outperforms the baselines by a large margin in all clustering metrics. By creating a state-utterance dictionary, we further demonstrate how we could augment original data by following the extracted structure. The extra training data is coherent logically but creates more variety in surface formats, thus provides a significant performance boost for end-to-end response generation. The proposed Multi-Response Data Augmentation (MRDA) beats recent work~\cite{gritta-etal-2021-conversation} using Most Frequent Sampling in a single-turn setting without annotated states. 
\section{Methodology}
\subsection{Problem Formulation}
We aim to discover a probabilistic semantic structure shared by dialogues from the same domain. We formulate the problem as labeling each dialogue with a sequence of dialogue states. A structure is then extracted by calculating the transition frequencies between pairs of states. Each conversational exchange $x_i$, a pair of system and user utterances at time step $i$, corresponds to a dialogue state $\mathbf{z}_i$, which tracks the status of the task and guide the upcoming dialogue. 

Commonly, dialogue states in task-oriented dialogue systems are defined as a set of slot-value pairs, which results in a huge amount of distinct states in total. To make the problem tractable, we count how many times each slot is modified without considering the actual slot values.
Specifically, 
\begin{equation}
    \mathbf{z}_i=[M(S_1), M(S_2),...,M(S_N)],
\end{equation}
where $S_j$ is a domain-specific slot, $M(S_j)$ is the number of changes of the slot $S_j$ from the beginning of the dialogue session, and $N$ is the number of slots in the given domain. Although it is hard to determine the number of states because the value of each slot could be updated for infinite times, it is reasonable to assume that the number of slots is available during inference. For example, by checking a few transcripts, a bot builder for the MultiWOZ \textit{attraction} domain can easily identify there are three slot types (name, type, area) that they need to fill in.

\subsection{Slot Boundary Detection and Clustering}
\begin{figure*}[t]
    \centering
    \includegraphics[width = 0.9\linewidth]{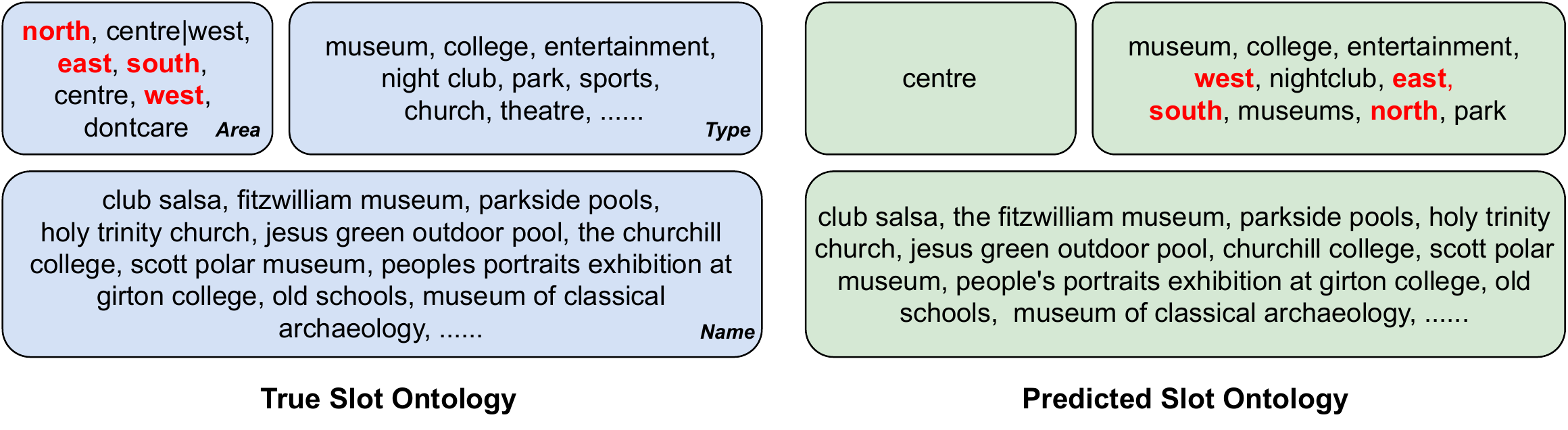}
    \caption{True slot ontology \textit{v.s.} predicted slot ontology of the \textit{attraction} domain in the MultiWOZ. Mis-clustered tokens are marked in \textbf{bold} and \textcolor{red}{red}. Slot names are unknown but it will not affect the structure extraction procedure.}
    \label{fig:sbd}
\end{figure*}
In a task-oriented dialogue system, slots are predefined in a domain ontology, and the system needs to identify their values to accomplish users' intents. For example, in order to book a taxi service, we need to fill the values of four slots: \textit{``leave-at'': 4 p.m., ``arrive-by'': 6 p.m., ``departure'': Palo Alto}, and \textit{``destination'': San Jose}. However, such a slot ontology is usually not available in a real scenario.
We thus define a preliminary sub-task of Slot Boundary Detection (SBD) and clustering for dialogue structure extraction. 
Given a target domain $G$, a set of dialogues $D$, and the number of slots $N$, the sub-task is to find all token spans that are possible slots in the domain $G$, and assign them into $N$ separate slot groups $\{S'_1, S'_2,...,S'_N\}$. As mentioned, we assume we do not have the slot ontology but $N$ is available.

For the SBD task, we retain the slot annotation in the conventional BIO scheme but drop the slot name labels. Table~\ref{tab:bio} shows examples in three task-oriented dialogue datasets. 
\begin{table}[ht]
\resizebox{\linewidth}{!}{
\begin{tabular}{ll}
\toprule
\multicolumn{2}{l}{\textbf{MultiWOZ}}                                                                              \\
\midrule
\multicolumn{1}{l|}{Utt.}        & {[}usr{]} a train to London King Cross that departs after 08:15 \\
\multicolumn{1}{l|}{Slots}      & O O O O B I I O O O B                                                    \\
\toprule
\multicolumn{2}{l}{\textbf{ATIS}}                                                                                  \\
\midrule
\multicolumn{1}{l|}{Utt.} & i want to fly from baltimore to dallas round trip                                \\
\multicolumn{1}{l|}{Slots}      & O O O O O B O B B I                                                              \\
\toprule
\multicolumn{2}{l}{\textbf{Snips}}                                                                                 \\
\midrule
\multicolumn{1}{l|}{Utt.} & book a restaurant for eight people in six years                                  \\
\multicolumn{1}{l|}{Slots}      & O O B O B O B I I                                                          \\
\bottomrule
\end{tabular}
}
\caption{Slot boundary annotation in the BIO scheme. Examples are from the MultiWOZ~\cite{budzianowski-etal-2018-multiwoz}, ATIS~\cite{tur2010left}, and Snips~\cite{coucke2018snips} datasets.}
\vspace{-2mm}
\label{tab:bio}
\end{table}

We hypothesize that the capability to identify slot tokens is transferable across domains. We train a BERT-based slot detector on some domains and apply it to an unseen domain. A \texttt{[CLS]} classification embedding is inserted as the first token and a \texttt{[SEP]} token is added as the final token. Given an input token sequence $\mathbf{x}=(x_1, ...,x_T)$, the final hidden states of BERT ($\textbf{h}_t$) is
fed into a softmax layer to classify over three labels: ``B'', ``I'', and ``O''. 

\begin{equation}
    y_t = \text{softmax}(\mathbf{W}\mathbf{h}_t + \mathbf{b}). 
\end{equation}
To make the process compatible with the BERT WordPiece tokenizer~\cite{wu2016google}, we assign the original label of a word to all its sub-tokens. This model is trained end-to-end to minimize with cross-entropy loss. For each token span $\mathbf{T}_i = [T_{i1},...,T_{ik}]$, if their slot labels predicted are $[\text{B}, \text{I},...,\text{I}] (k>1)$ or B$(k=1)$, and the label of token $T_{ik+1}$ is predicted as B or O, then $\mathbf{T}_i$ is considered as a slot token span. 

The pre-trained BERT model provides a powerful contextualized token representation. Therefore, we reuse the final hidden states for slot clustering. Mathematically, the token span $\mathbf{T}_i$ is encoded as
\begin{equation}
    \overline{\textbf{h}}_i=\frac{1}{k}\sum_{j=1}^{k}\mathbf{h}_{ij},
\end{equation}
where $\mathbf{h}_{i1},...,\mathbf{h}_{ik}$ are the final hidden states of $\mathbf{T}_i = [T_{i1},...,T_{ik}]$. The BERT representations are contextualized, so the same token spans appearing in different contexts have different encodings. By doing so, one token span can be assigned to multiple slot clusters simultaneously. For example, ``Palo Alto'' can be both a departure city and an arrival city, depending on its context.
By clustering the token span encodings, we can assign each of them into one of the $N$ groups and derive a fake slot ontology.
\begin{equation}
    S'_j = \text{clustering}(\overline{\textbf{h}}_i), j \in \{1,...,N\},
\end{equation}
where $S'_j$ is the $j$-th predicted slot group. Three clustering algorithms including KMeans~\cite{arthur2006k}, Birch~\cite{zhang1996birch}, and Agglomerative Clustering~\cite{mullner2011modern} are evaluated. Note that there is no guarantee that $S'_j$ can be mapped to any predefined slot type $S_j$. A clustered example is shown in Figure~\ref{fig:sbd}. More details about clustering are explained in section~\ref{sec:exp}.

\subsection{Deterministic Dialogue State Labeling}
\label{sec:det}
The slot boundary detection and clustering are followed by a deterministic procedure to construct the dialogue structure. To begin with, we initialize the dialogue state as $\mathbf{z}_0=[0,0,...,0]$. Then in dialogue turn $k$, for each slot token span $\mathbf{T}_i$ detected, if the clustering algorithm determines $\mathbf{T}_i \in S'_j$,  
we increment $M(S'_j)$ by one. 
Table~\ref{tab:attraction} demonstrates this procedure. In this way, we label each dialogue session with its extracted dialogue states without any state annotation. The dialogue structure is then depicted by representing distinct dialogue states as nodes. Due to the variety of $M(S'_j)$, the number of dialogue states is always larger than the number of slots, as shown in Table~\ref{tab:multiwoz_stats}. We connect an edge between a pair of nodes if there is such a transition in the data, and the edge is labeled as the normalized transition probability from the parent node. 
\begin{table}[h]
\centering
\resizebox{0.95\linewidth}{!}{
\begin{tabular}{l|ccccc}
\toprule
\textbf{Domain}  & Taxi & Restaurant & Hotel & Attraction & Train \\
\midrule
\textbf{\#samples} & 435  & 1,311      & 635   & 150        & 345   \\
\textbf{\#slots}   & 4    & 7          & 10    & 3          & 6     \\
\textbf{\#states}  & 29   & 206        & 734   & 11         & 85    \\
\bottomrule
\end{tabular}
}
\caption{Statistics of the MultiWOZ~\cite{budzianowski-etal-2018-multiwoz} dataset. \textbf{\#states} are number of annotated distinct dialogue states.}
\vspace{-2mm}
\label{tab:multiwoz_stats}
\end{table}

\section{Experiment}
\label{sec:exp}
\subsection{Datasets}
MultiWOZ~\cite{budzianowski-etal-2018-multiwoz} is a common benchmark for studying task-oriented dialogues. It has 8,420/1,000/1,000 dialogues for train, validation, and test, respectively. We use its revised version MultiWOZ 2.1~\cite{eric-etal-2020-multiwoz}, which has the same dialogue transcripts but with cleaner state label annotation. The MultiWOZ has five domains of dialogues: \textit{taxi, restaurant, hotel, attraction}, and \textit{train}. We hold out each of the domain for testing and use the remaining four domains for SBD training. Note that some of the target slots are not presented in the training slots, \textit{e.g.,} ``\textit{stay}'', ``\textit{stars}'', and ``\textit{internet}'' only appear in the \textit{hotel} domain.

To evaluate the transferability of the approach, we also tried to train the slot boundary detector on another two public datasets, ATIS~\cite{tur2010left, goo-etal-2018-slot} and Snips~\cite{coucke2018snips}. The ATIS dataset includes recordings of people making flight reservations and contains 4,478 utterances in its training set. The Snips dataset is collected from the Snips personal voice assistant and contains 13,084 training utterances. We train the SBD model on their training split and test on the selected domain of MultiWOZ. Examples are shown in Table~\ref{tab:bio}.
\begin{table*}[]
\centering
\resizebox{0.85\linewidth}{!}{
\begin{tabular}{r|ccccc|ccccc}
\toprule
                  & \multicolumn{5}{c|}{$\mathbf{F}1_{slot}$}                                                  & \multicolumn{5}{c}{$\mathbf{F}1_{token}$}                                                  \\
\midrule
Method            & Taxi          & Restaurant    & Hotel         & Attraction    & Train         & Taxi          & Restaurant    & Hotel         & Attraction    & Train         \\
\midrule
spaCy             & 0.43          & 0.48          & 0.47          & 0.33          & 0.39          & 0.28          & 0.21          & 0.21          & 0.16          & 0.23          \\
TOD-BERT\textsubscript{ATIS}     & 0.57          & 0.56          & 0.52          & 0.45          & 0.62          & 0.57          & 0.54          & 0.44          & 0.43          & 0.60          \\
TOD-BERT\textsubscript{SNIPS}    & 0.50          & 0.53          & 0.48          & 0.41          & 0.52          & 0.55          & 0.49          & 0.41          & 0.37          & 0.51          \\
TOD-BERT\textsubscript{MWOZ} & \textbf{0.90} & \textbf{0.89} & \textbf{0.84} & \textbf{0.91} & \textbf{0.84} & \textbf{0.90} & \textbf{0.89} & \textbf{0.82} & \textbf{0.91} & \textbf{0.84} \\
\bottomrule
\end{tabular}
}
\caption{Slot boundary detection results tested in the MultiWOZ.}
\vspace{-2mm}
\label{tab:sbd}
\end{table*}

\begin{table*}[t]
\centering
\resizebox{0.9\linewidth}{!}{
\begin{tabular}{l|ccccc|ccccc|ccccc}
\toprule
               & \multicolumn{5}{c}{ARI}                                                       & \multicolumn{5}{c}{AMI}                                                       & \multicolumn{5}{c}{SC}                                                        \\
\midrule
               & Taxi          & Rest.         & Hotel         & Attr.         & Train         & Taxi          & Rest.         & Hotel         & Attr.         & Train         & Taxi          & Rest.         & Hotel         & Attr.         & Train         \\
\midrule
Random         & 0.00          & 0.00          & 0.00          & 0.00          & 0.00          & 0.00          & 0.00          & 0.00          & 0.00          & 0.00          & -             & -             & -             & -             & -             \\
VRNN    & 0.05          & 0.00          & 0.00          & 0.00          & 0.00          & 0.05          & 0.02          & 0.00          & 0.01          & 0.06          & -          & -          & -          & -          & -          \\
BERT-KMeans    & 0.02          & 0.01          & 0.01          & 0.01          & 0.01          & 0.11          & 0.09          & 0.02          & 0.03          & 0.06          & 0.11          & 0.08          & 0.06          & 0.13          & 0.09          \\
TOD-BERT-mlm   & 0.02          & 0.01          & 0.01          & 0.03          & 0.02          & 0.13          & 0.11          & 0.03          & 0.06          & 0.10          & 0.12          & 0.08          & 0.06          & 0.17          & 0.09          \\
TOD-BERT-jnt   & 0.03          & 0.02          & 0.02          & 0.03          & 0.03          & 0.16          & 0.13          & 0.06          & 0.08          & 0.14          & 0.09          & 0.08          & 0.06          & 0.13          & 0.07          \\
BERT-spaCy     & 0.01          & 0.06          & 0.04          & 0.01          & 0.01          & 0.09          & 0.18          & 0.12          & 0.06          & 0.08          & -             & -             & -             & -             & -             \\
TOD-BERT-spaCy & 0.01          & 0.03          & 0.05          & 0.02          & 0.01          & 0.09          & 0.15          & 0.12          & 0.05          & 0.05          & -             & -             & -             & -             & -             \\
TOD-BERT-SBD\textsubscript{MWOZ}   & \textbf{0.15}          & 0.00          & 0.00          & 0.00         & 0.05          & 0.17          & 0.13          & 0.04          & 0.06          & 0.16          & \textbf{0.39} & \textbf{0.34} & \textbf{0.27} & \textbf{0.44} & \textbf{0.34} \\
TOD-BERT-DET\textsubscript{ATIS}  & 0.08          & 0.05          & 0.09          & 0.03          & 0.06          & 0.26          & 0.22          & 0.25          & 0.15          & 0.26          & -             & -             & -             & -             & -             \\
TOD-BERT-DET\textsubscript{SNIPS} & 0.06          & 0.05          & 0.11          & 0.03          & 0.04          & 0.25          & 0.23          & 0.22          & 0.09          & 0.22          & -             & -             & -             & -             & -             \\
TOD-BERT-DET\textsubscript{MWOZ}  & \textbf{0.15} & \textbf{0.22} & \textbf{0.24} & \textbf{0.33} & \textbf{0.24} & \textbf{0.39} & \textbf{0.48} & \textbf{0.44} & \textbf{0.44} & \textbf{0.44} & -             & -             & -             & -             & -  \\
\bottomrule
\end{tabular}
}
\caption{Structure extraction results using clustering metrics in the MultiWOZ dataset. SC is omitted for methods that do not encode utterances directly. Results using BERT-Birch and BERT-Agg are reported in Appendix~\ref{sec:appendix}.}
\vspace{-2mm}
\label{tab:struc-extract}
\end{table*}

\subsection{Setup}
We conduct extensive experiments to compare our approach with different baseline models. The ground truth construction follows the same deterministic procedure by counting the modification times of annotated slot values, instead of the spans predicted by our algorithm. We describe details of the baseline models as follows.
\begin{itemize}[leftmargin=*]
    \item \textbf{Random} Every conversational turn is randomly assigned a state by selecting a number from 1 to the ground truth \#states in Table~\ref{tab:multiwoz_stats}. 
    
    \item \textbf{VRNN} Dialogues are reconstructed with Variational Recurrent Neural Networks~\cite{shi-etal-2019-unsupervised, qiu-etal-2020-structured}, which is a recurrent version of Variational Auto-Encoder (VAE). The extracted structure represents the transition among discrete latent variables.
    
    \item \textbf{BERT-KMeans/Birch/Agg} Each conversational turn is encoded by BERT with the final hidden state of \texttt{[CLS]} token. The utterance encodings are then clustered with Kmeans, Birch, and Aggolomerative clustering methods. 
    \begin{equation}
        \mathbf{z} = \text{clustering}(\mathbf{h}_{\text{CLS}})
    \end{equation}
    Number of clusters are directly set to the \#states.
    
    \item \textbf{TOD-BERT-mlm/jnt} This is similar to the previous baseline but encoding utterances with TOD-BERT. TOD-BERT~\cite{wu-etal-2020-tod} is based on BERT architecture and trained on nine task-oriented datasets using two loss functions: Masked Language Modeling (MLM) loss and Response Contrastive Loss (RCL). TOD-BERT-mlm only uses the MLM loss, while TOD-BERT-jnt is jointly trained with both loss functions. The utterance encodings are clustered with KMeans.
    
    \item \textbf{(TOD-)BERT-spaCy} Instead of training a slot boundary detector based on BERT, we implement a heuristic-based detector with spaCy\footnote{\url{https://spacy.io/}}. Words are labeled as slot spans if they are nouns. Suppose it detects $n$ slot words $\{w_1...,w_n\}$ in the $u_i$ utterance, the $j$-th word has $|w_j|$ sub-tokens, the BERT/TOD-BERT encoding of the $k$-th sub-token of this word is $\mathbf{h}_{jk}$. Then we represent this turn as:
    \begin{equation}
        \mathbf{u_i} = \frac{1}{n}\sum_{j=1}^{n}\frac{1}{|w_j|}\sum_{k=1}^{|w_i|}\mathbf{h}_{jk}.
    \end{equation}
    In this method, we do not cluster slot representations, but we use average slot embeddings to represent the whole utterance. Then $\mathbf{u_i}$ are clustered to \#states clusters with KMeans: 
    \begin{equation}
        \mathbf{z_i} = \text{clustering}(\mathbf{u_i}).
    \end{equation}

    \item \textbf{TOD-BERT-SBD}\textsubscript{MWOZ} This is similar to the previous approach. But instead of using a heuristic-based detector, the TOD-BERT is trained for SBD in training domains of MultiWOZ and detect slot tokens in the test domain, and then we use those detected slot embeddings to represent each utterance.

    \item \textbf{TOD-BERT-DET}\textsubscript{ATIS/SNIPS/MWOZ} The TOD-BERT is trained for SBD in the ATIS, Snips, or the MultiWOZ training domains. Then in the test domain of MultiWOZ, we follow the deterministic dialogue state labeling process described in section~\ref{sec:det}, instead of clustering utterance embeddings, to extract a structure.
\end{itemize}
We use English uncased BERT-Base model, which has 12 layers, 12 heads, and 768 hidden states. We train BERT (or TOD-BERT) on the Slot Boundary Detection (SBD) task with AdamW~\cite{loshchilov2017decoupled} optimizer using a dropout rate of 0.1. The model is trained with an initial learning rate of $5\mathrm{e}{-5}$ for 5 epochs on two NVIDIA Tesla V100 GPUs.

\subsection{Results and Analysis}
Table~\ref{tab:sbd} shows the empirical results of Slot Boundary Detection. We report the $\mathbf{F}1$ score in both slot level ($\mathbf{F}1_{slot}$) and token level ($\mathbf{F}1_{token}$). In the slot level, a slot prediction is considered correct only when an exact match is found, which doesn't reward token overlap (partial match). In general, BERT-based slot boundary detectors perform better than the heuristic-based detector. Because utterances in MultiWOZ share similar interaction behaviors and utterance lengths, it makes the model easier to transfer from one domain to another within MultiWOZ than from the ATIS and Snips to the MultiWOZ.

We further analyze the performance of structure extraction, as shown in Table~\ref{tab:struc-extract}. We evaluate the model performance with clustering metrics, testing whether utterances assigned to the same state are more similar than utterances of different states. Given the knowledge of the ground truth dialogue state assignments and the model assignments of the same utterances, the Rand Index (RI) is a function that measures the similarity of the two assignments. Mathematically, 
\begin{equation}
    \textbf{RI} = \frac{a+b}{C_2^{n_{\text{samples}}}}, \\
\end{equation}
where $a$ is the number of pairs of elements that are assigned to the same set by both the ground truth and the model, $b$ is the number of pairs of elements that are assigned to different sets by both, $C_2^{n_{\text{samples}}}$ is the total number of pairs in the dataset. The Adjusted Rand Index (ARI) corrects for chance and guarantees that random assignments have an ARI close to 0. For a comprehensive analysis, we also report Adjusted Mutual Information (AMI) and Silhouette Coefficient (SC). While both ARI and AMI require the knowledge of the ground truth classes, the Silhouette Coefficient (SC) evaluates the model itself but the computation needs utterance representations. Thus, we do not report SC for methods such as TOD-BERT-DET.

\begin{figure}[th]
    \centering
    \includegraphics[width = \linewidth]{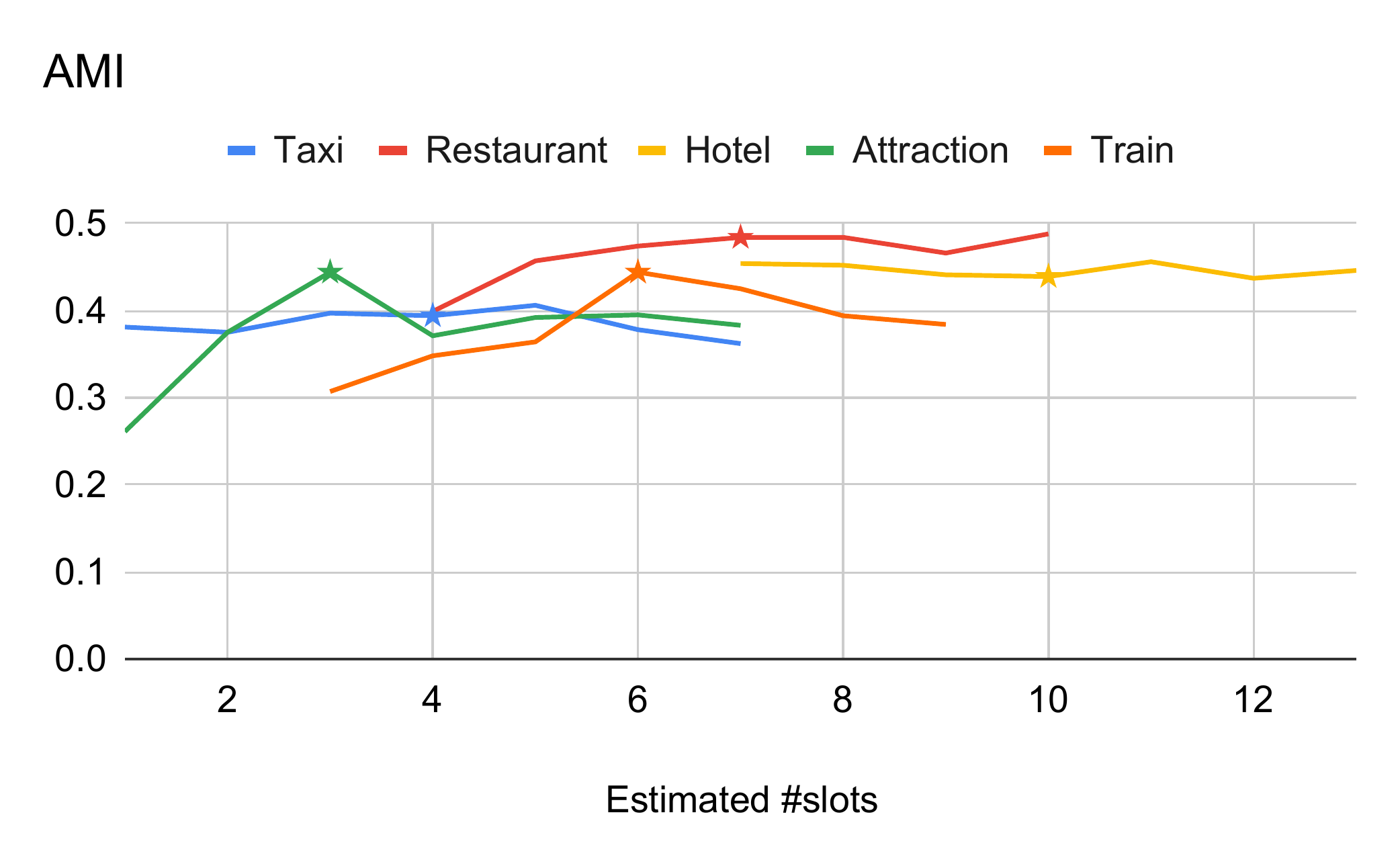}
    \caption{Evaluation of the proposed TOD-BERT-DET\textsubscript{MWOZ}'s robustness to estimated \#slots. Stars are the ground truth. ARI results are attached in Appendix~\ref{sec:appendix}.}
    \vspace{-2mm}
    \label{fig:slot_robustness}
\end{figure}

We observe a negligible effect of using different clustering algorithms on the structure extraction performance. As we can see in Table~\ref{tab:struc-extract}, the VRNN baseline performs not so well, because their dialogue states are defined in a latent space while the ground truth we compare with is based on the accumulative status of slots. Switching the encoder from the original BERT to TOD-BERT provides a slight improvement. Using a spaCy-based detector can have inaccurate slot detection, so the performance of (TOD-)BERT-spaCy are worse than TOD-BERT-SBD\textsubscript{MWOZ}. Simply averaging the detected slot token encodings for utterance clustering will also lose the information of individual slot changes. Compared with these baselines, our approach TOD-BERT-DET\textsubscript{MWOZ/ATIS/SNIPS} based on slot boundary detection and deterministic dialogue state labeling outperforms others by a large margin. In Figure~\ref{fig:slot_robustness}, we show the robustness of the proposed TOD-BERT-DET\textsubscript{MWOZ} to an inaccurate estimation of \#slots. In Appendix~\ref{sec:appendix}, we show example utterances that are predicted as the same state in different domains.

\section{Data Augmentation}
Conversations have an intrinsic one-to-many property, meaning that multiple responses can be appropriate for the same dialog context~\cite{zhang2020task}.
Leveraging this property, we augmented training data to improve end-to-end dialogue response generation based on the extracted structure. Specifically, we build a dictionary mapping from the dialogue state to its different valid utterances. Then we enable this dictionary to create additional data during training, which allows a language model to learn a balanced distribution. 
In the following sections, we will briefly introduce the task of single-turn dialogue response generation, the baseline augmentation approach called Most Frequent Sampling~\cite{gritta-etal-2021-conversation}, and the proposed Multi-Response Data Augmentation.

\begin{figure*}[ht!] 
  \begin{minipage}{0.19\linewidth} 
    \includegraphics[width=1.0\textwidth]{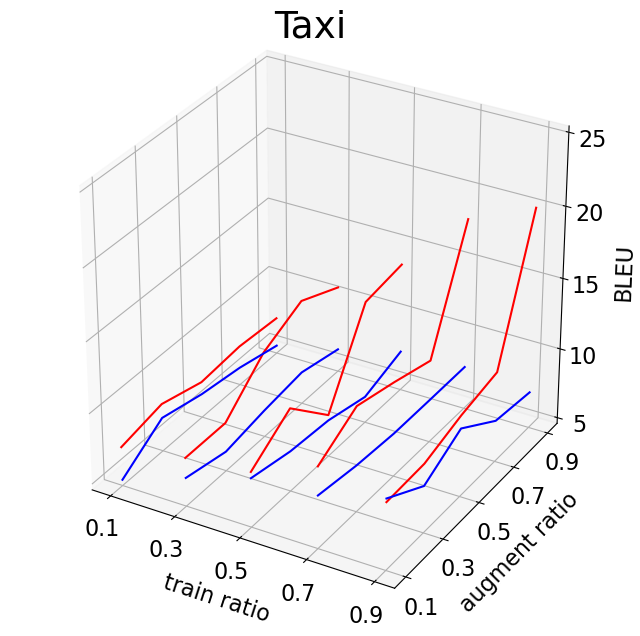}
  \end{minipage}
  \begin{minipage}{0.19\linewidth} 
    \includegraphics[width=1.0\textwidth]{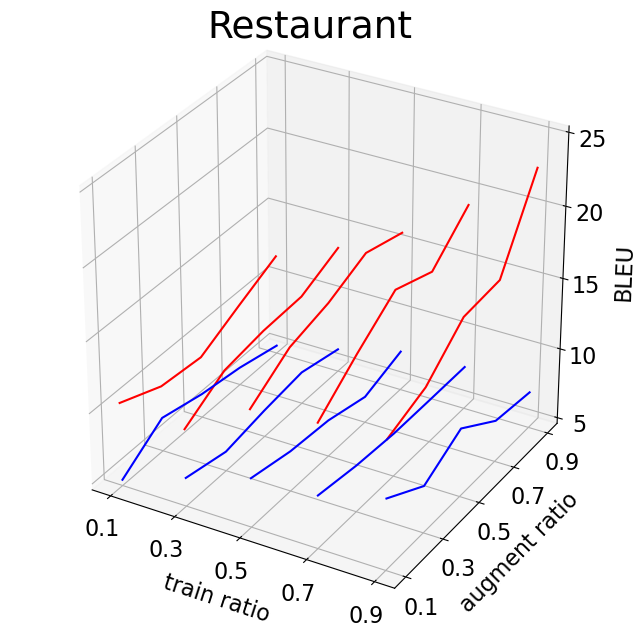}
  \end{minipage}
  \begin{minipage}{0.19\linewidth} 
    \includegraphics[width=1.0\textwidth]{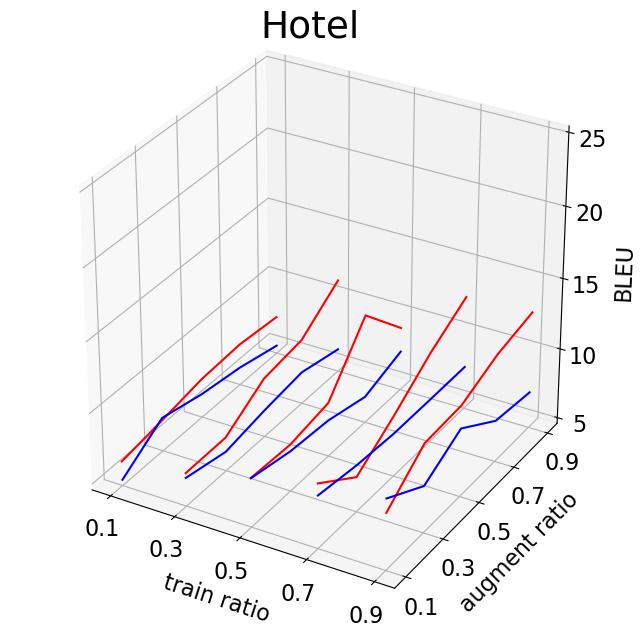}
  \end{minipage}
  \begin{minipage}{0.19\linewidth} 
    \includegraphics[width=1.0\textwidth]{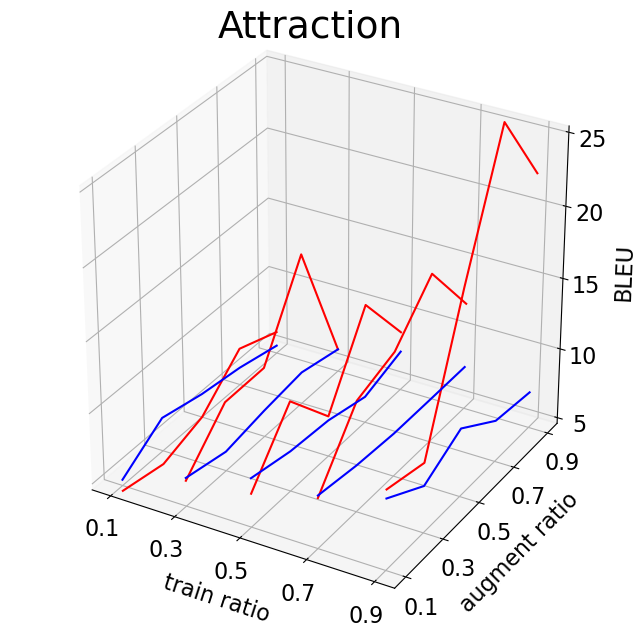}
  \end{minipage}
  \begin{minipage}{0.19\linewidth} 
    \includegraphics[width=1.0\textwidth]{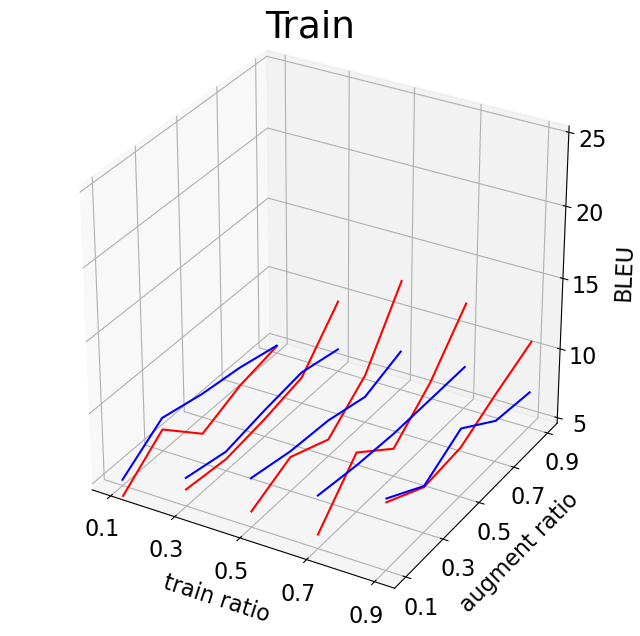}
  \end{minipage}
  \caption{Data Augmentation (BLEU$\uparrow$) in the MultiWOZ. \textcolor{blue}{Blue}: MFS. \textcolor{red}{Red}: MRDA (ours).} 
  \vspace{-2mm}
  \label{fig:aug_bleu}
\end{figure*}
\subsection{Single-Turn Dialogue Generation}
Training a single-turn dialogue response generative model is to learn an autoregressive (AR) model that maximize the log-likelihood $\mathcal{L}$ of ground truth response $R=x_{n+1},...,x_T$ conditioned on dialogue history $C=x_1,...,x_n$, which is encoded by dialogue state $\mathbf{z}$:
\begin{equation}
    \begin{aligned}
        \mathcal{L} &= \sum_{i \in D}\log P(R_i | C_i)  \\
                    &= \sum_{i \in D}\log \prod_{t=n+1}^{T}p(x_t|x_1,...,x_{t-1}),
    \end{aligned}
\end{equation}
where $i$ is each turn in dialogue corpus $D$. For a number of dialogue history $C_i$ belonging to the same state $\mathbf{z}$, there exits $K$ different system responses $R^{(1)},...,R^{(K)}$ that are valid, \textit{i.e.,} for $j=1,...,K, \exists i \in D$ \textit{s.t.} $(\mathbf{z}_i, R_i)=(\mathbf{z}, R^{(j)})$. We denote the valid system response set for dialogue state $\mathbf{z}$ as $\mathcal{V}(\mathbf{z})$.

\subsection{Most Frequent Sampling}
\citet{gritta-etal-2021-conversation} proposed Most Frequent Sampling (MFS) as a data augmentation strategy based on the annotated conversational graph. MFS generates novel training instances so that the most frequent agent actions are preceded by new histories, which is one or more original paths leading to common actions. The authors observed the best performance when they combined extra data augmented from MFS with the baseline training data.

\subsection{Multi-Response Data Augmentation}
However, augmented with the Most Frequent Sampling, it may exaggerate the frequency imbalance among valid responses, resulting in a lower response diversity. The original MFS also depends on annotated dialogue states from the MultiWOZ. To alleviate the problems, we propose Multi-Response Data Augmentation (MRDA) to balance the valid response distribution of each state $\mathbf{z}$ based on our extracted dialogue structure. Concretely, for each dialog turn $i$ with state-response pair $(\mathbf{z}_i, R_i)$, we incorporate other valid system responses under the same state, \textit{i.e.,} $R_{i'}, i' \neq i$ with $\mathbf{z}_{i'} = \mathbf{z}_i$, as additional training data for turn $i$. The new objective function becomes:
\begin{equation}
    \mathcal{L_{\text{aug}}} = \sum_{i \in D} \sum_{R_{i'} \in \mathcal{V}^*(\mathbf{z}_i)}  \log P(R_{i'} | C_i),
\end{equation}
where $\mathcal{V}^*(\mathbf{z}_i) \subseteq \mathcal{V}(\mathbf{z}_i)$ is a subset of the valid response set $\mathcal{V}(\mathbf{z}_i)$ of dialogue state $\mathbf{z}_i$, $\mathbf{z}_i$ is the predicted dialogue state of history $C_i$. The idea is similar to the Multi-Action Data Augmentation (MADA) proposed by~\citet{zhang2020task}, but our method doesn't need to train an action decoder to generate responses and no annotated dialogue states are required.

\subsection{Setup}
We compare our MRDA approach with the MFS baseline in the MultiWOZ dataset. We used the ground truth dialogue states for MFS as in its original paper. For MRDA, we hold out each of the domains for testing and use the remaining four domains for SBD training and dialogue state prediction. The data of each held-out domain is split into train (60\%), valid (20\%), and test (20\%) for the language model training and testing. To evaluate both methods in a realistic setting where training data is limited and augmentation is required, we adjust the ratio between actually used training data and total training data, denoted by $r_{\text{train}}$. Moreover, to explore the impact of augmented data size, we define $r_{\text{aug}}$ as the ratio between the size of augmented samples and used training samples. The \textsc{DialoGPT}~\cite{zhang-etal-2020-dialogpt} model is trained with the data for 5 epochs with a learning rate of $3\mathrm{e}{-5}$ to generate single-turn responses. 

\subsection{Results and Analysis}
The generation perplexity and BLEU scores in the five domains of the MultiWOZ are reported in Table~\ref{tab:aug}. Both augmentation methods first double the original training samples, \textit{i.e.,} $r_{\text{train}}=1.0,r_{\text{aug}}=1.0$. By augmenting the data, we reduce the perplexity by an average of 1.24 and improve the BLEU score by an average of 12.01. The results also demonstrate our approach outperforms the MFS baseline by an average of 2.14 in perplexity and 12.37 in BLEU, because the MRDA balances the valid response distribution. Our approach also doesn't require any annotation of the test domain.
\begin{table}[t]
\centering
\resizebox{0.9\linewidth}{!}{
\begin{tabular}{l|ccccc}
\toprule
\textbf{Perplexity$\downarrow$}         & Taxi  & Rest. & Hotel & Attr. & Train \\
\midrule
Original train     & 4.88  & 4.46       & 6.16  & 7.75       & 6.58  \\
+ MFS     & 5.34  & 6.54       & 7.17  & 8.39       & 6.91  \\
+ MRDA & \textbf{4.64}  & \textbf{3.69}       & \textbf{5.57}  & \textbf{3.91}       & \textbf{5.83}  \\
\midrule
\midrule
\textbf{BLEU$\uparrow$}               & Taxi  & Rest. & Hotel & Attr. & Train \\
\midrule
Original train     & 9.88  & 12.54       & 8.68  & 8.46       & 8.93  \\
+ MFS     & 9.73  & 12.56       & 7.54  & 9.21       & 7.64  \\
+ MRDA & \textbf{22.13} & \textbf{18.77}      & \textbf{10.66} & \textbf{47.79}      & \textbf{9.20} \\
\bottomrule
\end{tabular}
}
\caption{Response generation in the MultiWOZ with data augmentation ($r_{\text{train}}=1.0,r_{\text{aug}}=1.0 $).}
\vspace{-4mm}
\label{tab:aug}
\end{table}

To explore the impact of available training data size and augmented data size, we try different combinations of the $r_{\text{train}}$ and $r_{\text{aug}}$, and illustrate the results in Figure~\ref{fig:aug_bleu} (numbers attached in Appendix~\ref{sec:appendix}). The figure shows that: \textit{(i)} Our MRDA approach constantly improves the generation performance, and it outperforms the MFS baseline regardless of the original data size. \textit{(ii)}  Data augmentation based on a larger training set provides more performance boost because the language model is trained with more data and different valid responses are balanced. These observations suggest that our extracted dialogue structure can successfully augment meaningful dialogue for response generation, with the potential to improve other dialogue downstream tasks such as policy learning and summarization. We also include example augmented dialogues in the Appendix~\ref{sec:appendix}.

Table~\ref{tab:overlap} reports how many states are overlapped in the MultiWOZ, using the slot value annotation and our dialogue state definition. It shows that our test set has no distinct dialogue state that never appears in the train or valid sets, while this may not be the case in practice. The MRDA method creates new instances that follow existing dialogue flows but with different surface formats, while it remains a compelling direction to create completely new state sequences by discovering causal dependencies in the extracted structures.

\begin{table}[h!]
\centering
\resizebox{0.9\linewidth}{!}{
\begin{tabular}{lccccc}
\toprule
\textbf{State Overlap}       & Taxi & Rest. & Hotel & Attr. & Train \\
\midrule
Train Only         & 9    & 97         & 387   & 0          & 25    \\
Valid Only         & 0    & 12         & 51    & 0          & 7     \\
\textbf{Test Only}          & \textbf{0}    & \textbf{0}          & \textbf{0}     & \textbf{0}          & \textbf{0}     \\
Train \& Valid     & 18   & 68         & 148   & 10         & 38    \\
Train \& Test      & 16   & 55         & 99    & 9          & 32    \\
Test \& Valid      & 16   & 62         & 141   & 9          & 37    \\
Train \& Valid \& Test & 16   & 55         & 99    & 9          & 32    \\
\bottomrule
\end{tabular}
}
\caption{Annotated dialogue state overlap across train, valid, and test splits in the MultiWOZ dataset.}
\vspace{-4mm}
\label{tab:overlap}
\end{table}

\section{Related Works}
Extensive works have been done on studying the structures of dialogues. \citet{jurafsky1997switchboard} learned semantic structures based on human annotations. While such annotations are expensive and vary in quality, recent research shifted their focus to unsupervised approaches. By reconstructing the original dialogues with discrete latent variable models, we can extract a structure representing the transition among the variables. In this direction, people have tried Hidden Markov Models~\cite{chotimongkol2008learning, ritter-etal-2010-unsupervised, zhai-williams-2014-discovering}, Variational Auto-Encoders (VAEs)~\cite{kingma2013auto}, and its recurrent version Variational Recurrent Neural Networks (VRNNs)~\cite{chung2015recurrent, shi-etal-2019-unsupervised, qiu-etal-2020-structured}. More recently, \citet{sun2021unsupervised} proposed an Edge-Enhanced Graph Auto-Encoder (EGAE) architecture to model local-contextual and global structural information. Meanwhile, \citet{xu-etal-2021-discovering} integrates Graph Neural Networks into a Discrete Variational Auto-Encoder to discover structures in open-domain dialogues. However, it is hard to interpret or evaluate the extracted structure because of the implicitly defined latent variables. 

Benefiting from the pre-training technique~\cite{mccann2017learned, howard-ruder-2018-universal, peters-etal-2018-deep, devlin-etal-2019-bert}, the Transformer architecture~\cite{vaswani2017attention} can be trained on generic corpora and adapted to specific downstream tasks.
In dialogue systems, \citet{wu-etal-2020-tod} pre-trained the BERT model~\cite{devlin-etal-2019-bert} on task-oriented dialogues for intent recognition, dialogue state tracking, dialogue act prediction, and response selection. ~\citet{peng2021soloist, hosseini2020simple} parameterize classical modular task-oriented dialogue system with an autoregressive language model GPT-2~\cite{radford2019language}. \textsc{DialoGPT}~\cite{zhang-etal-2020-dialogpt} extends the GPT-2 to conversational response generation in single-turn dialogue settings. 
In this work, we demonstrate how we can adapt a pre-trained Transformer for structure extraction in task-oriented dialogues. Our approach of detecting slot boundaries first is also related to the work of~\citet{hudecek-etal-2021-discovering} but uses a different model.

The extracted structures are proved useful in multiple downstream tasks. \citet{xu-etal-2021-discovering} use the structure to guide coherent dialogue generation in open domains. 
\citet{shi-etal-2019-unsupervised} and~\citet{zhao-etal-2019-rethinking} use the learned structures for dialogue policy learning. \citet{gritta-etal-2021-conversation} augment training data with the proposed Most Frequent Sampling (MFS) to improve the success rate of task-oriented dialog systems. 
\citet{zhang2020task} propose a Multi-Action Data Augmentation (MADA) framework guiding the dialog policy to learn a balanced action distribution. Nevertheless, both MFS and MADA are based on annotated dialogue states. Our work shows that the extracted structure can also be leveraged for data augmentation and alternative sampling strategies could be used. 
    

\section{Conclusion \& Future Work}
This paper proposes a simple yet effective approach for structure extraction in task-oriented dialogues. We define a task of Slot Boundary Detection and clustering to approximate the dialogue ontology. 
We extract a semantic structure that explicitly depicts the state transitions in task-oriented dialogues, without using state annotation during inference. Extensive experiments demonstrate that our approach is superior to the baseline models in all the domains of the MultiWOZ dataset. In addition, we demonstrate how to augment dialogue data based on our extracted structures to improve end-to-end response generation remarkably. Comprehensive analysis and downstream application study of these extracted dialogue structures are two main challenges in future work.


\bibliography{anthology,custom}
\bibliographystyle{acl_natbib}

\appendix
\section{Appendix}
\label{sec:appendix}

\begin{figure}[ht!] 
  \begin{minipage}{0.49\linewidth} 
    \includegraphics[width=1.0\textwidth]{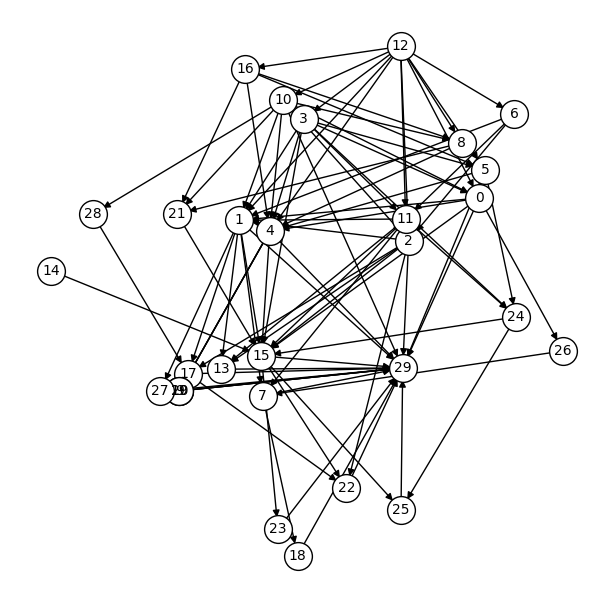}
  \end{minipage}
  \begin{minipage}{0.49\linewidth} 
    \includegraphics[width=1.0\textwidth]{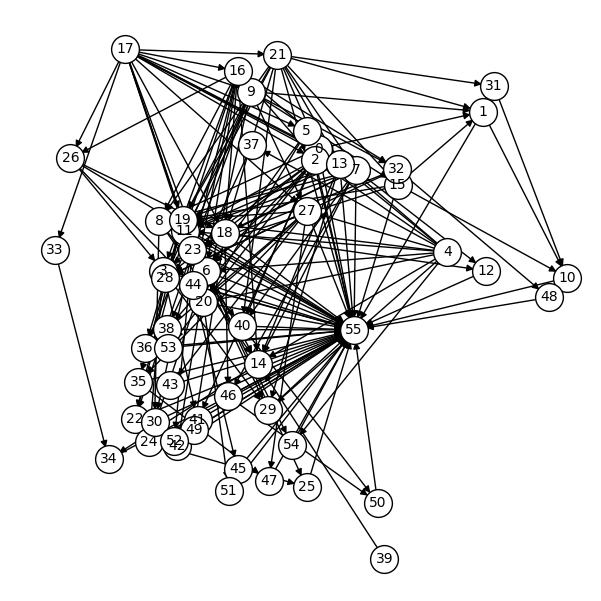}
  \end{minipage}
  \caption{Dialogue structure in the \textit{taxi} domain of the MultiWOZ. The structure on the left is from annotated dialogue states, while the right one is extracted by our approach.} 
  \label{fig:taxi}
\end{figure}

\begin{figure}[ht!] 
  \begin{minipage}{0.49\linewidth} 
    \includegraphics[width=1.0\textwidth]{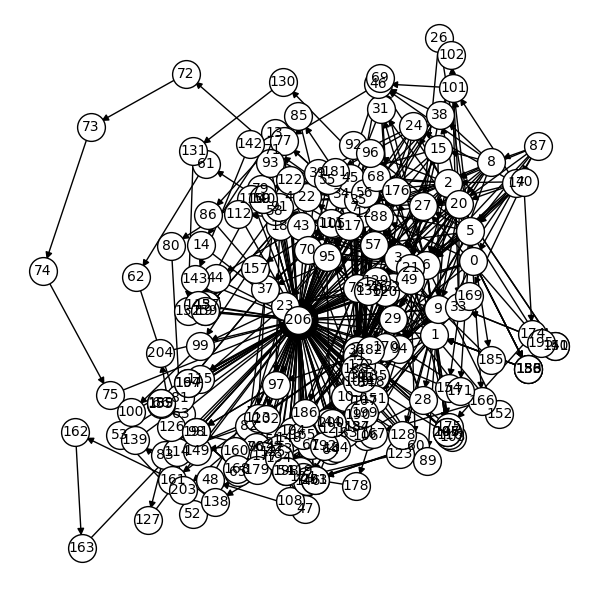}
  \end{minipage}
  \begin{minipage}{0.49\linewidth} 
    \includegraphics[width=1.0\textwidth]{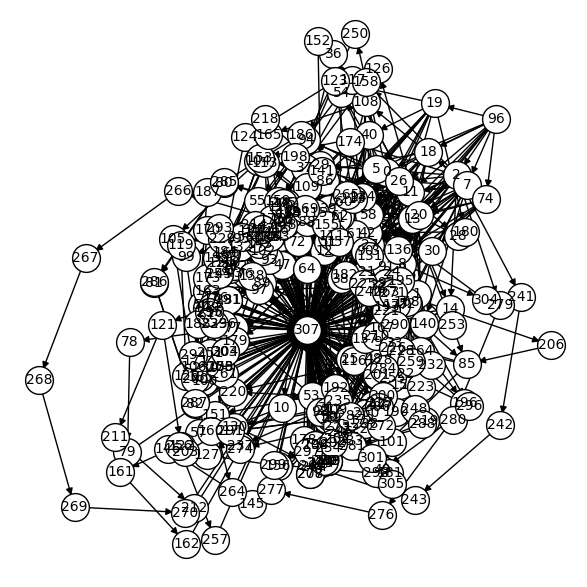}
  \end{minipage}
  \caption{Dialogue structure in the \textit{restaurant} domain of the MultiWOZ. The structure on the left is from annotated dialogue states, while the right one is extracted by our approach.} 
  \label{fig:rest}
\end{figure}

\begin{figure}[ht!] 
  \begin{minipage}{0.49\linewidth} 
    \includegraphics[width=1.0\textwidth]{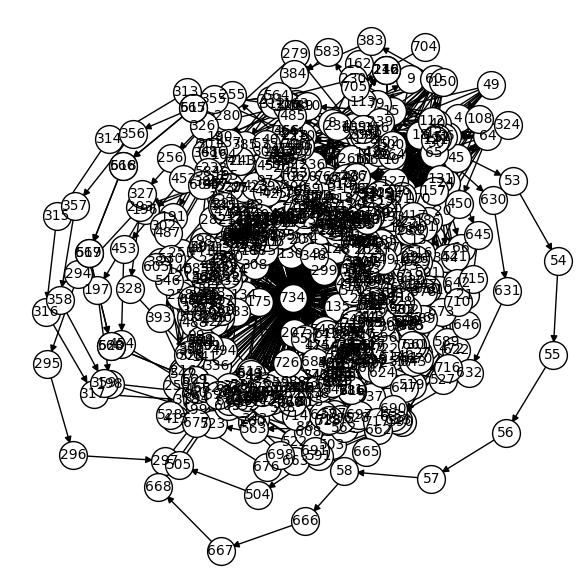}
  \end{minipage}
  \begin{minipage}{0.49\linewidth} 
    \includegraphics[width=1.0\textwidth]{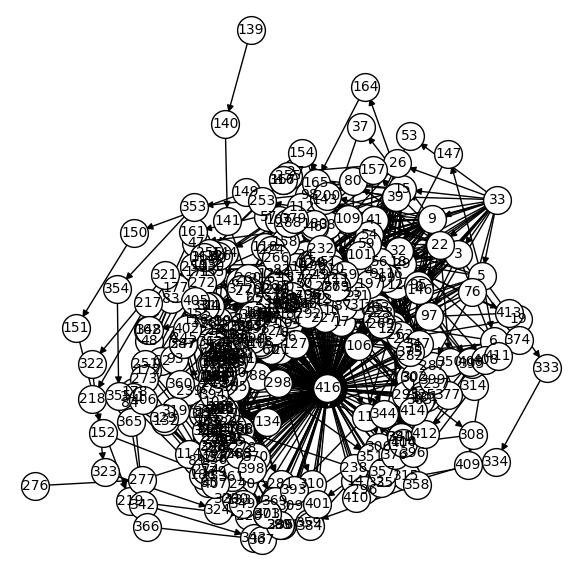}
  \end{minipage}
  \caption{Dialogue structure in the \textit{hotel} domain of the MultiWOZ. The structure on the left is from annotated dialogue states, while the right one is extracted by our approach.} 
  \label{fig:hotel}
\end{figure}

\begin{figure}[ht!] 
  \begin{minipage}{0.49\linewidth} 
    \includegraphics[width=1.0\textwidth]{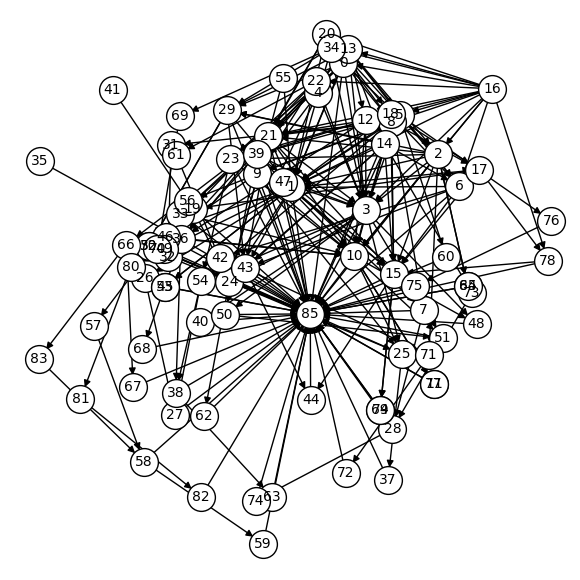}
  \end{minipage}
  \begin{minipage}{0.49\linewidth} 
    \includegraphics[width=1.0\textwidth]{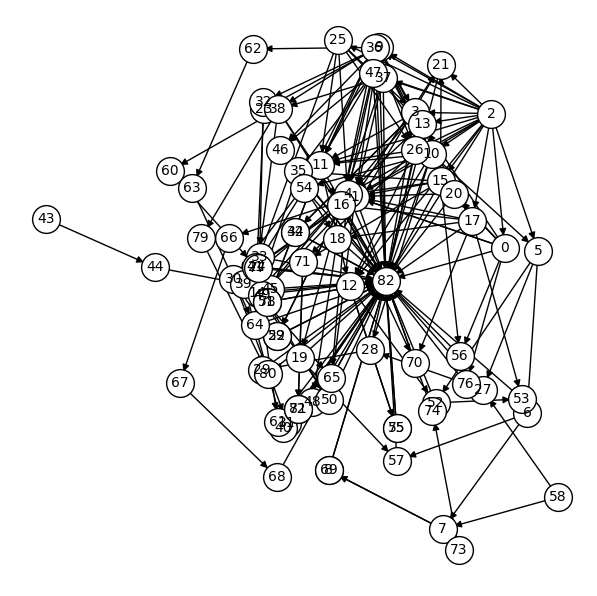}
  \end{minipage}
  \caption{Dialogue structure in the \textit{train} domain of the MultiWOZ. The structure on the left is from annotated dialogue states, while the right one is extracted by our approach.} 
  \label{fig:train}
\end{figure}

\begin{figure}[th]
    \centering
    \includegraphics[width = \linewidth]{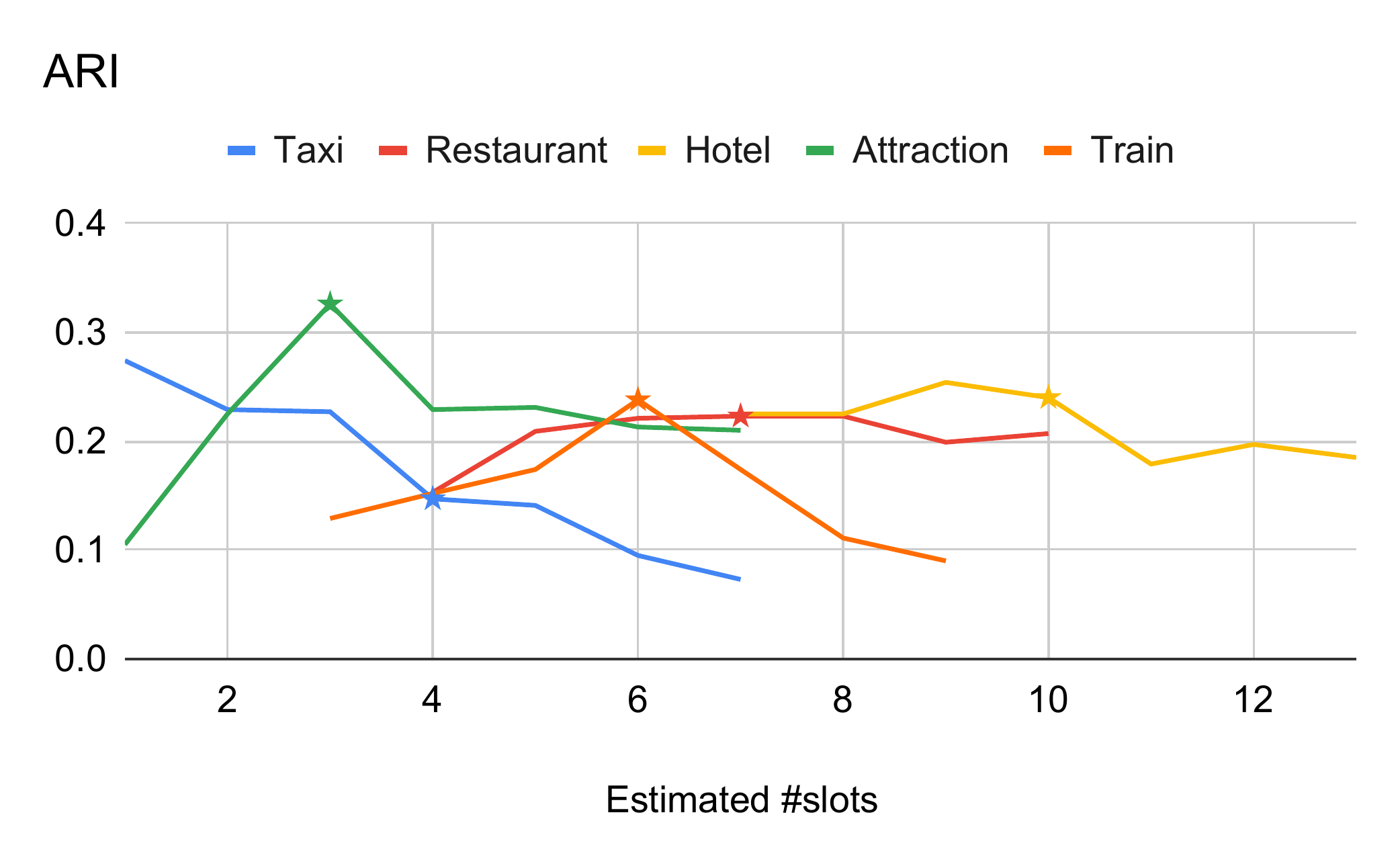}
    \caption{Evaluation of the proposed TOD-BERT-DET\textsubscript{MWOZ}'s robustness to estimated \#slots. Stars are the ground truth.}
\end{figure}

\begin{figure*}[t] 
  \begin{minipage}{0.19\linewidth} 
    \includegraphics[width=1.0\textwidth]{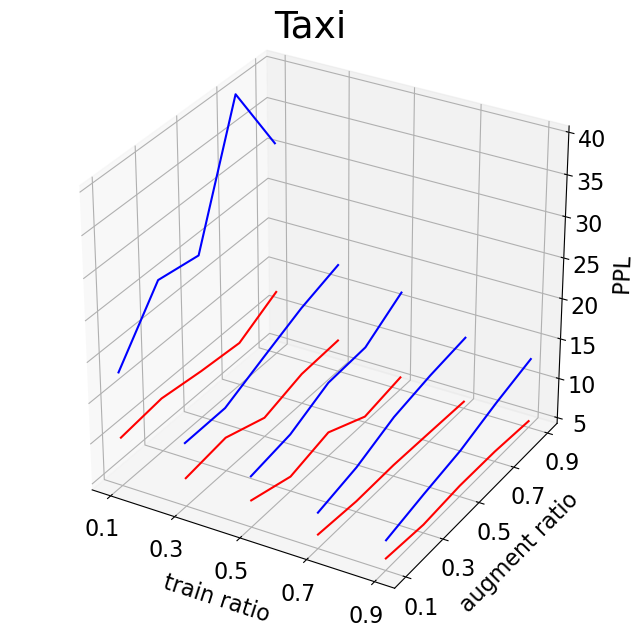}
  \end{minipage}
  \begin{minipage}{0.19\linewidth} 
    \includegraphics[width=1.0\textwidth]{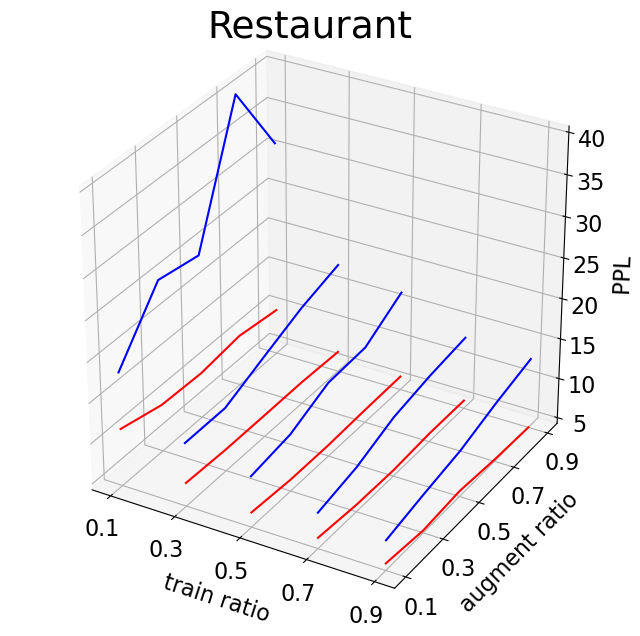}
  \end{minipage}
  \begin{minipage}{0.19\linewidth} 
    \includegraphics[width=1.0\textwidth]{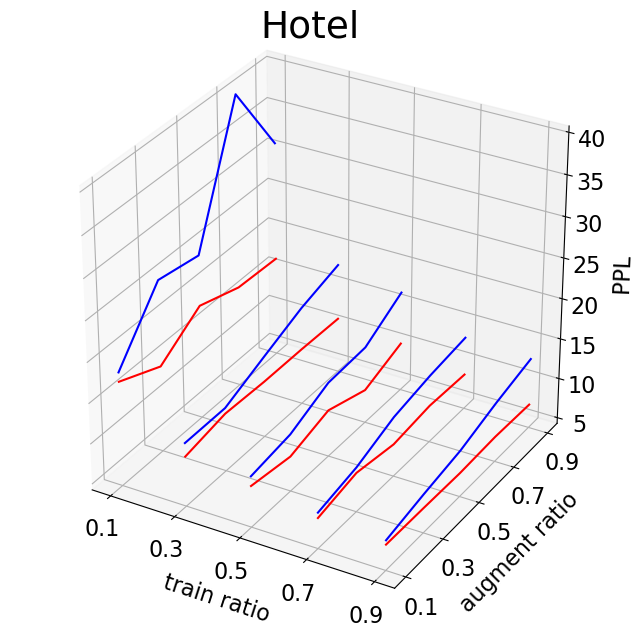}
  \end{minipage}
  \begin{minipage}{0.19\linewidth} 
    \includegraphics[width=1.0\textwidth]{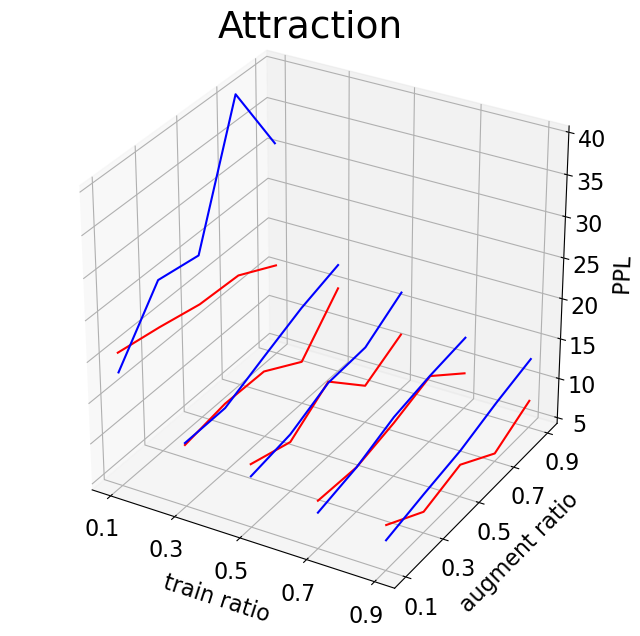}
  \end{minipage}
  \begin{minipage}{0.19\linewidth} 
    \includegraphics[width=1.0\textwidth]{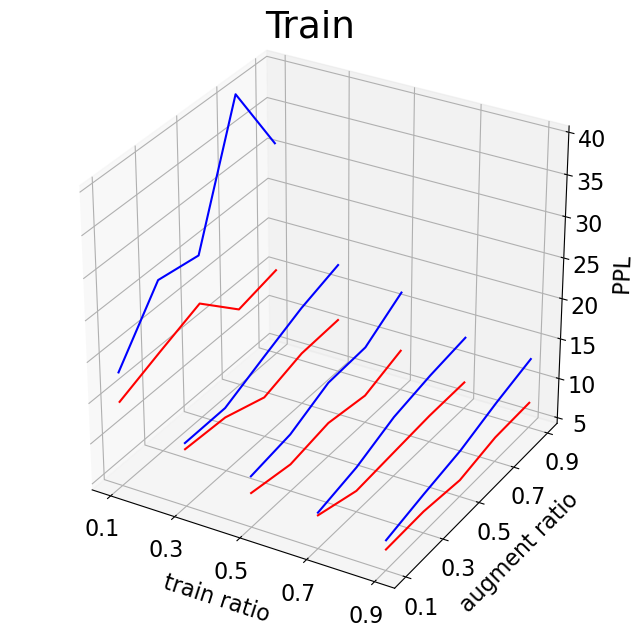}
  \end{minipage}
  \caption{Data Augmentation (perplexity$\downarrow$) in the MultiWOZ. \textcolor{blue}{Blue}: MFS. \textcolor{red}{Red}: MRDA (ours).} 
  \vspace{-2mm}
  \label{fig:aug_ppl}
\end{figure*}

\begin{table*}[]
\centering
\resizebox{0.7\linewidth}{!}{
\begin{tabular}{c|ccccc}
\toprule
$r_{\text{train}}|r_{\text{aug}}$ & 0.1          & 0.3          & 0.5          & 0.7         & 0.9          \\
\midrule
0.1       & 10.28 (11.96) & 10.90 (14.07) & 10.17 (13.68) & 9.67  (15.39) & 12.25 (14.42) \\
0.3       & 7.77 (8.48)   & 8.44 (9.17)   & 6.66 (9.16)   & 8.05 (9.89)  & 8.27 (9.45)   \\
0.5       & 7.60 (7.87)   & 6.09 (8.34)   & 7.28 (8.05)   & 5.00 (9.01)  & 5.85 (8.56)   \\
0.7       & 6.01 (6.65)   & 5.58 (6.61)   & 5.65 (6.83)   & 5.40 (7.96)  & 5.13 (7.03)   \\
0.9       & 5.75 (6.17)   & 5.28 (6.79)   & 5.62 (6.51)   & 5.52 (7.18)  & 5.08 (7.14)   \\
\bottomrule
\end{tabular}
}
\caption{Data Augmentation with MRDA (perplexity$\downarrow$) in the \textbf{Taxi} domain of the MultiWOZ. Numbers in the parenthesis are using MFS. }
\label{tab:my-table}
\end{table*}

\begin{table*}[]
\centering
\resizebox{0.7\linewidth}{!}{
\begin{tabular}{c|ccccc}
\toprule
$r_{\text{train}}|r_{\text{aug}}$ & 0.1          & 0.3          & 0.5          & 0.7         & 0.9          \\
\midrule
0.1             & 11.36 (10.39) & 10.09 (12.18) & 9.91 (14.97) & 10.57 (14.53) & 9.90 (14.81) \\
0.3             & 7.17 (7.65)   & 6.96 (8.64)   & 6.96 (9.34)  & 6.99 (8.26)   & 6.78 (8.04)  \\
0.5             & 6.10 (6.47)   & 5.70 (6.78)   & 5.66 (7.27)  & 5.85 (7.07)   & 5.96 (7.93)  \\
0.7             & 5.61 (5.84)   & 5.13 (5.86)   & 5.00 (6.64)  & 5.29 (6.19)   & 5.27 (6.46)  \\
0.9             & 5.14 (5.32)   & 4.58 (5.64)   & 4.96 (5.78)  & 4.50 (5.75)   & 4.28 (6.18)    \\
\bottomrule
\end{tabular}
}
\caption{Data Augmentation with MRDA (perplexity$\downarrow$) in the \textbf{Restaurant} domain of the MultiWOZ. Numbers in the parenthesis are using MFS. }
\label{tab:my-table}
\end{table*}

\begin{table*}[]
\centering
\resizebox{0.7\linewidth}{!}{
\begin{tabular}{c|ccccc}
\toprule
$r_{\text{train}}|r_{\text{aug}}$ & 0.1          & 0.3          & 0.5          & 0.7         & 0.9          \\
\midrule
0.1        & 17.17 (18.31) & 14.90 (25.47) & 18.39 (24.60) & 16.75 (40.37) & 16.52 (30.99) \\
0.3        & 10.45 (12.12) & 11.45 (12.15) & 11.22 (14.33) & 11.21 (16.43) & 11.08 (17.94) \\
0.5        & 9.38 (10.57)  & 8.60 (11.35)  & 10.06 (13.52) & 8.40 (13.81)  & 10.22 (16.71) \\
0.7        & 8.05 (8.71)   & 9.12 (9.80)   & 8.28 (11.69)  & 8.85 (12.66)  & 8.62 (13.31)  \\
0.9        & 7.45 (7.98)   & 7.36 (9.08)   & 7.22 (10.04)  & 7.39 (11.65)  & 7.20 (12.98)     \\
\bottomrule
\end{tabular}
}
\caption{Data Augmentation with MRDA (perplexity$\downarrow$) in the \textbf{Hotel} domain of the MultiWOZ. Numbers in the parenthesis are using MFS. }
\label{tab:my-table}
\end{table*}

\begin{table*}[]
\centering
\resizebox{0.7\linewidth}{!}{
\begin{tabular}{c|ccccc}
\toprule
$r_{\text{train}}|r_{\text{aug}}$ & 0.1          & 0.3          & 0.5          & 0.7         & 0.9          \\
\midrule
0.1             & 20.70 (21.61) & 19.74 (21.30) & 18.57 (30.25) & 18.26 (23.63) & 15.67 (30.17) \\
0.3             & 11.88 (14.29) & 12.69 (15.65) & 12.52 (16.07) & 9.61 (17.93)  & 14.98 (18.71) \\
0.5             & 12.06 (12.18) & 10.41 (14.43) & 13.66 (13.49) & 8.95 (14.35)  & 11.37 (15.27) \\
0.7             & 10.17 (10.66) & 9.81 (11.70)  & 10.97 (11.74) & 12.59 (11.27) & 8.78 (13.61)  \\
0.9             & 9.84 (10.00)  & 6.89 (10.46)  & 8.26 (12.42)  & 5.28 (11.52)  & 7.68 (11.20)    \\
\bottomrule
\end{tabular}
}
\caption{Data Augmentation with MRDA (perplexity$\downarrow$) in the \textbf{Attraction} domain of the MultiWOZ. Numbers in the parenthesis are using MFS. }
\label{tab:my-table}
\end{table*}

\begin{table*}[]
\centering
\resizebox{0.7\linewidth}{!}{
\begin{tabular}{c|ccccc}
\toprule
$r_{\text{train}}|r_{\text{aug}}$ & 0.1          & 0.3          & 0.5          & 0.7         & 0.9          \\
\midrule
0.1        & 14.71 (18.41) & 16.74 (18.31) & 18.67 (22.44) & 13.96 (21.89) & 15.05 (28.61) \\
0.3        & 11.36 (11.29) & 10.98 (14.00) & 9.25 (14.07)  & 10.68 (15.73) & 10.92 (13.93) \\
0.5        & 8.51 (9.90)   & 7.62 (9.26)   & 8.47 (10.11)  & 7.70 (11.13)  & 9.30 (12.30)  \\
0.7        & 8.38 (8.49)   & 6.88 (9.46)   & 7.24 (10.28)  & 7.58 (10.78)  & 7.61 (10.18)  \\
0.9        & 6.85 (8.30)   & 6.93 (8.80)   & 6.33 (8.70)   & 7.31 (8.42)   & 7.43 (9.16)      \\
\bottomrule
\end{tabular}
}
\caption{Data Augmentation with MRDA (perplexity$\downarrow$) in the \textbf{Train} domain of the MultiWOZ. Numbers in the parenthesis are using MFS. }
\label{tab:my-table}
\end{table*}

\begin{table*}[]
\centering
\resizebox{0.7\linewidth}{!}{
\begin{tabular}{c|ccccc}
\toprule
$r_{\text{train}}|r_{\text{aug}}$ & 0.1          & 0.3          & 0.5          & 0.7         & 0.9          \\
\midrule
0.1        & 7.34 (6.69)  & 7.97 (5.73)  & 7.15 (6.18)  & 7.41 (5.57)  & 7.19 (5.67)  \\
0.3        & 8.02 (7.79)  & 8.02 (7.75)  & 10.67 (7.05) & 12.04 (7.51) & 10.77 (7.11) \\
0.5        & 8.49 (7.04)  & 10.47 (8.77) & 7.55 (6.96)  & 13.27 (5.86) & 13.70 (4.63) \\
0.7        & 10.34 (8.34) & 12.03 (7.77) & 11.29 (5.60) & 10.44 (4.97) & 18.15 (6.39) \\
0.9        & 9.35 (7.79)  & 9.46 (8.93)  & 10.36 (6.58) & 10.98 (7.20) & 20.16 (8.31)      \\
\bottomrule
\end{tabular}
}
\caption{Data Augmentation with MRDA (BLEU$\uparrow$) in the \textbf{Taxi} domain of the MultiWOZ. Numbers in the parenthesis are using MFS. }
\label{tab:my-table}
\end{table*}

\begin{table*}[]
\centering
\resizebox{0.7\linewidth}{!}{
\begin{tabular}{c|ccccc}
\toprule
$r_{\text{train}}|r_{\text{aug}}$ & 0.1          & 0.3          & 0.5          & 0.7         & 0.9          \\
\midrule
0.1              & 10.47 (9.45)  & 9.24 (9.32)   & 8.96 (9.30)   & 10.39 (6.20)  & 11.76 (9.44)  \\
0.3              & 10.04 (9.98)  & 11.74 (10.33) & 12.20 (10.57) & 12.35 (11.29) & 13.63 (11.32) \\
0.5              & 12.84 (12.14) & 14.70 (11.89) & 15.46 (13.45) & 16.71 (12.78) & 15.96 (13.58) \\
0.7              & 13.33 (12.33) & 15.61 (13.29) & 17.67 (12.49) & 16.68 (12.80) & 19.13 (12.13) \\
0.9              & 13.64 (13.84) & 14.71 (11.99) & 17.13 (12.93) & 17.39 (13.03) & 22.87 (13.66)     \\
\bottomrule
\end{tabular}
}
\caption{Data Augmentation with MRDA (BLEU$\uparrow$) in the \textbf{Restaurant} domain of the MultiWOZ. Numbers in the parenthesis are using MFS. }
\label{tab:my-table}
\end{table*}

\begin{table*}[]
\centering
\resizebox{0.7\linewidth}{!}{
\begin{tabular}{c|ccccc}
\toprule
$r_{\text{train}}|r_{\text{aug}}$ & 0.1          & 0.3          & 0.5          & 0.7         & 0.9          \\
\midrule
0.1         & 6.32 (5.00) & 6.80 (6.97)  & 7.35 (6.27)  & 7.54 (5.84)  & 7.28 (5.14)  \\
0.3         & 6.94 (6.60) & 6.98 (5.96)  & 8.79 (6.48)  & 9.22 (6.86)  & 11.28 (6.21) \\
0.5         & 8.07 (8.05) & 7.91 (7.43)  & 8.43 (7.17)  & 12.32 (6.45) & 9.12 (7.39)  \\
0.7         & 9.16 (8.32) & 7.05 (7.88)  & 8.92 (7.64)  & 11.00 (7.62) & 12.67 (7.62) \\
0.9         & 8.61 (9.60) & 10.89 (7.90) & 11.06 (9.43) & 12.24 (7.52) & 12.89 (7.15)      \\
\bottomrule
\end{tabular}
}
\caption{Data Augmentation with MRDA (BLEU$\uparrow$) in the \textbf{Hotel} domain of the MultiWOZ. Numbers in the parenthesis are using MFS. }
\label{tab:my-table}
\end{table*}

\begin{table*}[]
\centering
\resizebox{0.7\linewidth}{!}{
\begin{tabular}{c|ccccc}
\toprule
$r_{\text{train}}|r_{\text{aug}}$ & 0.1          & 0.3          & 0.5          & 0.7         & 0.9          \\
\midrule
0.1              & 4.20 (4.36)  & 3.61 (2.59)  & 4.60 (4.60)  & 7.24 (3.15)  & 6.16 (5.52)   \\
0.3              & 6.40 (5.26)  & 9.50 (9.40)  & 9.55 (6.60)  & 15.35 (7.39) & 6.21 (4.17)   \\
0.5              & 6.96 (5.90)  & 10.96 (5.77) & 7.46 (7.36)  & 13.06 (7.25) & 8.81 (7.45)   \\
0.7              & 8.14 (4.93)  & 12.40 (7.65) & 13.41 (7.76) & 16.54 (2.58) & 12.19 (7.14)  \\
0.9              & 10.22 (6.94) & 9.52 (8.06)  & 19.13 (6.51) & 27.91 (8.71) & 22.50 (10.01)      \\
\bottomrule
\end{tabular}
}
\caption{Data Augmentation with MRDA (BLEU$\uparrow$) in the \textbf{Attraction} domain of the MultiWOZ. Numbers in the parenthesis are using MFS. }
\label{tab:my-table}
\end{table*}

\begin{table*}[]
\centering
\resizebox{0.7\linewidth}{!}{
\begin{tabular}{c|ccccc}
\toprule
$r_{\text{train}}|r_{\text{aug}}$ & 0.1          & 0.3          & 0.5          & 0.7         & 0.9          \\
\midrule
0.1         & 3.83 (3.77) & 6.13 (3.72) & 3.35 (4.14) & 4.51 (2.94) & 5.09 (3.31)  \\
0.3         & 5.78 (5.66) & 5.42 (5.33) & 5.84 (3.55) & 6.48 (4.17) & 9.73 (4.88)  \\
0.5         & 5.70 (6.24) & 7.02 (6.69) & 5.79 (4.67) & 8.02 (6.66) & 12.52 (5.21) \\
0.7         & 5.58 (3.81) & 8.79 (6.83) & 6.56 (5.44) & 8.85 (6.34) & 12.21 (7.43) \\
0.9         & 9.34 (6.45) & 7.82 (6.61) & 8.08 (4.25) & 9.51 (6.84) & 10.79 (6.70)      \\
\bottomrule
\end{tabular}
}
\caption{Data Augmentation with MRDA (BLEU$\uparrow$) in the \textbf{Train} domain of the MultiWOZ. Numbers in the parenthesis are using MFS. }
\label{tab:my-table}
\end{table*}

\begin{table*}[]
\centering
\resizebox{0.9\linewidth}{!}{
\begin{tabular}{l|ccccc|ccccc|ccccc}
\toprule
               & \multicolumn{5}{c}{ARI}                                                       & \multicolumn{5}{c}{AMI}                                                       & \multicolumn{5}{c}{SC}                                                        \\
\midrule
               & Taxi          & Rest.         & Hotel         & Attr.         & Train         & Taxi          & Rest.         & Hotel         & Attr.         & Train         & Taxi          & Rest.         & Hotel         & Attr.         & Train         \\
\midrule
Random         & 0.00          & 0.00          & 0.00          & 0.00          & 0.00          & 0.00          & 0.00          & 0.00          & 0.00          & 0.00          & -             & -             & -             & -             & -             \\
VRNN    & 0.05          & 0.00          & 0.00          & 0.00          & 0.00          & 0.05          & 0.02          & 0.00          & 0.01          & 0.06          & -          & -          & -          & -          & -          \\
BERT-KMeans    & 0.02          & 0.01          & 0.01          & 0.01          & 0.01          & 0.11          & 0.09          & 0.02          & 0.03          & 0.06          & 0.11          & 0.08          & 0.06          & 0.13          & 0.09          \\
BERT-Birch     & 0.02          & 0.01          & 0.01          & 0.01          & 0.01          & 0.12          & 0.09          & 0.02          & 0.03          & 0.06          & 0.07          & 0.06          & 0.08          & 0.09          & 0.07          \\
BERT-Agg       & 0.02          & 0.01          & 0.01          & 0.01          & 0.01          & 0.11          & 0.09          & 0.02          & 0.05          & 0.07          & 0.08          & 0.05          & 0.08          & 0.11          & 0.07          \\
TOD-BERT-mlm   & 0.02          & 0.01          & 0.01          & 0.03          & 0.02          & 0.13          & 0.11          & 0.03          & 0.06          & 0.10          & 0.12          & 0.08          & 0.06          & 0.17          & 0.09          \\
TOD-BERT-jnt   & 0.03          & 0.02          & 0.02          & 0.03          & 0.03          & 0.16          & 0.13          & 0.06          & 0.08          & 0.14          & 0.09          & 0.08          & 0.06          & 0.13          & 0.07          \\
BERT-spaCy     & 0.01          & 0.06          & 0.04          & 0.01          & 0.01          & 0.09          & 0.18          & 0.12          & 0.06          & 0.08          & -             & -             & -             & -             & -             \\
TOD-BERT-spaCy & 0.01          & 0.03          & 0.05          & 0.02          & 0.01          & 0.09          & 0.15          & 0.12          & 0.05          & 0.05          & -             & -             & -             & -             & -             \\
TOD-BERT-SBD\textsubscript{MWOZ}   & \textbf{0.15}          & 0.00          & 0.00          & 0.00         & 0.05          & 0.17          & 0.13          & 0.04          & 0.06          & 0.16          & \textbf{0.39} & \textbf{0.34} & \textbf{0.27} & \textbf{0.44} & \textbf{0.34} \\
TOD-BERT-DET\textsubscript{ATIS}  & 0.08          & 0.05          & 0.09          & 0.03          & 0.06          & 0.26          & 0.22          & 0.25          & 0.15          & 0.26          & -             & -             & -             & -             & -             \\
TOD-BERT-DET\textsubscript{SNIPS} & 0.06          & 0.05          & 0.11          & 0.03          & 0.04          & 0.25          & 0.23          & 0.22          & 0.09          & 0.22          & -             & -             & -             & -             & -             \\
TOD-BERT-DET\textsubscript{MWOZ}  & \textbf{0.15} & \textbf{0.22} & \textbf{0.24} & \textbf{0.33} & \textbf{0.24} & \textbf{0.39} & \textbf{0.48} & \textbf{0.44} & \textbf{0.44} & \textbf{0.44} & -             & -             & -             & -             & -  \\
\bottomrule
\end{tabular}
}
\caption{Complete structure extraction results using clustering metrics in the MultiWOZ dataset. SC is omitted for methods that do not encode utterances directly. 
BERT with different clustering methods are included.}
\label{tab:struc-extract-complete}
\end{table*}

\begin{table*}[]
\centering
\resizebox{\linewidth}{!}{
\begin{tabular}{c|l|l}
\toprule
\textbf{Domain}     & Original                                                                                                                                                                                                                                                                                                                                                                                                                                                                                                                                                                                                                                                                                                                                                                                                                                                                              & Augmented                                                                                                                                                                                                                                                                                                                                                                                                                                                                                                                                                                                                                                                                                                                                                                                                                                                        \\
\midrule
\textbf{Taxi}       & \begin{tabular}[c]{@{}l@{}}{[}USR{]} I would like a taxi from Saint John's college to Pizza Hut Fen Ditton. \\ {[}SYS{]} What time do you want to leave and what time do you want to arrive by?\\ {[}USR{]} I want to leave after 17:15. \\ {[}USR{]} Booking completed! \\ Your taxi will be blue honda Contact number is 07218068540\\ {[}USR{]} Thank you for all the help! I appreciate it.\\ {[}SYS{]} You are welcome.  Is there anything else I can help you with today?\\ {[}USR{]} No, I am all set.  Have a nice day. Bye.\\ {[}SYS{]} you too! thank you\end{tabular}                                                                                                                                                                                                                                                                                                      & \begin{tabular}[c]{@{}l@{}}{[}USR{]} From museum of classical archaeology to london kings cross \\train station, please\\ {[}SYS{]} I have ordered one for you their pick up time is 1:00pm. \\ Is this time okay or do you need a different time?\\ {[}USR{]} Please pick me up at the scott polar museum\\ {[}SYS{]} Okay, I've booked a taxi for you.\\ {[}USR{]} Could you give me the car type and contact number?\\ {[}SYS{]} Certainly, the car is a white Honda, and the contact number is 07135860962.\\ {[}USR{]} Got it. Thanks for helping me.\\ {[}SYS{]} Your welcome. Have a great time.\end{tabular}                                                                                                                                                                                                                                               \\
\midrule
\textbf{Restaurant} & \begin{tabular}[c]{@{}l@{}}{[}USR{]} Can you help me find Restaurant One Seven?\\ {[}SYS{]} I have Restaurant One Seven located in Centre, \\phone number is 01223337766. \\ Would you like for me to make a reservation for you?\\ {[}USR{]} I would like to book for two on Wednesday, at 17:15.\\ {[}SYS{]} I have you booked Wednesday at 17:15 a table for 2, \\ this table will hold for 15 mins and your reference number is GXO64LG8.\\ {[}USR{]} Great, thanks!  That's all I needed!\\ {[}SYS{]} You're welcome! Have a great day. Goodbye.\end{tabular}                                                                                                                                                                                                                                                                                                                      & \begin{tabular}[c]{@{}l@{}}{[}USR{]} I am looking for a particular restaurant. It is called pizza hut city centre.\\ {[}SYS{]} It's a cheap italian place in the centre at Regent Street City Centre cb21ab. \\ The phone number is 01223323737. Would you like more info?\\ {[}USR{]} That sounds like just the place! \\ Can you book me a table for 2 at 19:45 on Thursday?\\ {[}SYS{]} You're all set. I've got a table for 2 at 19:45 this Thursday. \\ Your reference number is F3K2PQZZ.\\ {[}USR{]} Ok sounds good. Thank you!\\ {[}SYS{]} You are welcome. Is there anything else I can assist you with today?\\ {[}USR{]} That is all. Thank you for your help!\\ {[}SYS{]} You're welcome. Have a great day.\end{tabular}                                                                                                                             \\
\midrule
\textbf{Hotel}      & \begin{tabular}[c]{@{}l@{}}{[}USR{]} I need a place to stay in the north please.\\ {[}SYS{]} There are 13 places to stay in the north. \\ Do you have any other criteria?\\ {[}USR{]} I don't want to have to pay for parking.\\ {[}SYS{]} I have 5 different gueshouses listed. \\For what day and for how many please?\\ {[}USR{]} For 3 people starting on Wednesday and staying 2 nights.\\ {[}SYS{]} There are 11 hotels available actually. \\ Would you like to narrow it to a specific price range or star rating?\\ {[}USR{]} Just a guesthouse in the north with free parking. \\ I'll need a reference number then too please.\\ {[}SYS{]} I was able to get you in at Archway House. \\ Your reference number is WUMUZ0JX. \\ Can I help you with anything else?\\ {[}USR{]} That's all I need today. Thanks! Bye!\\ {[}SYS{]} Thank you for using our system!\end{tabular} & \begin{tabular}[c]{@{}l@{}}{[}USR{]} north area of town\\ {[}SYS{]} There are two hotels in that area with free parking: \\Ashley Hotel and the Lovell lodge.  \\ Would you like to book either of those?\\ {[}USR{]} Actually, I'd like a guest house.\\ {[}SYS{]} how about hamilton lodge? It sounds lovely.\\ {[}USR{]} Can you see if either of them are available \\for 2 nights for 5 people starting on Thursday, \\ I have no preference for which one. \\ {[}SYS{]} Both of those hotels are completely booked up when trying to fit that criteria. \\ Perhaps we can try different specifications?\end{tabular}                                                                                                                                                                                                                                           \\
\midrule
\textbf{Attraction} & \begin{tabular}[c]{@{}l@{}}{[}USR{]} Someone told me about a place called Old Schools. \\ It's supposed to be some kind of attraction. \\ Can you tell me anything about it?\\ {[}SYS{]} Old schools is an architecture museum in the centre area.  \\ The admission is free. Would you like tickets?\\ {[}USR{]} Yes, please. That would be great.\\ {[}SYS{]} I just checked and it is free. \\ The address is trinity lane and the phone number is 01223332320.  \\ Can I help you with anything else?\\ {[}USR{]} No, that is all for today. Thanks so much!\\ {[}SYS{]} You're welcome! Have a great day!\end{tabular}                                                                                                                                                                                                                                                           & \begin{tabular}[c]{@{}l@{}}{[}USR{]} Yes, hello. I'm looking for a particular attraction called Downing College.\\ {[}SYS{]} t is located on \textbackslash{}tregent street, in the centre.\\ {[}USR{]} Thank you so much!\\ {[}SYS{]} You are welcome! Did you need anything else today?\\ {[}USR{]} No, that was all. Good bye.\\ {[}SYS{]} Thanks and enjoy!\end{tabular}                                                                                                                                                                                                                                                                                                                                                                                                                                                                                     \\
\midrule
\textbf{Train}      & \begin{tabular}[c]{@{}l@{}}{[}USR{]} am looking for a train, it should go to cambridge and \\should depart from norwich\\ {[}SYS{]} What time and day are you looking to travel?\\ {[}USR{]} Yes, I would like travel on Monday and I would need to arrive by 08:30.\\ {[}SYS{]} TrainID TR9020 leaves norwich on monday and \\arrives in cambridge at 06:35.\\ {[}USR{]} What is the price for the train?\\ {[}SYS{]} The cost will be 17.60 pounds. \\ Would you like more information about that?\\ {[}USR{]} What time does the train leave?\\ {[}SYS{]} it leaves at 05:16",\\ {[}USR{]} Thanks, that's all. Good bye.\textbackslash{}n\\ {[}SYS{]} Goodbye and thanks for using the help desk.\end{tabular}                                                                                                                                                                         & \begin{tabular}[c]{@{}l@{}}{[}USR{]} I am looking for a train out of kings lynn please.\\ {[}SYS{]} Where will you be going?\\ {[}USR{]} I want to go to Cambridge.\\ {[}SYS{]} When will you be leaving?\\ {[}USR{]} yes. i  should leave after 13:45 and should leave on thursday\\ {[}SYS{]} I have a train leaving Thursday at 13:59, \\would you like me to book it for you?\\ {[}USR{]} Could you tell me when the train arrives in London Liverpool Street?\\ {[}SYS{]} That train arrives at 15:27. Would you like me to book it for you?\\ {[}USR{]} yes pliz.may i also get the arrival time\\ {[}SYS{]} The arrival time is by 15:27, do you want to book a seat?\\ {[}USR{]} Not yet. I just needed to get the details. Thanks for helping me. Goodbye.\\ {[}SYS{]} Thank you for using our services. Do you need any further assistance?\end{tabular} \\
\bottomrule

\end{tabular}
}
\caption{Examples of generated dialogues by the Multi-Response Data Augmentation in the MultiWOZ.}
\label{tab:my-table}
\end{table*}

\begin{table*}[ht!]
\centering
\resizebox{\linewidth}{!}{
\begin{tabular}{|l|l|l|l|}
\hline
State/Utts & State 0                                                                                                                                                                                                    & State 1                                                                                                                                                                                                                                                                & ... \\ \hline
\multicolumn{4}{|c|}{\textbf{Taxi}}\\ \hline
Utt. 0           & \begin{tabular}[c]{@{}l@{}}{[}usr{]} I would like a taxi from Saint John's college to Pizza Hut Fen Ditton.\\ {[}sys{]} What time do you want to leave and what time do you want to arrive by?\end{tabular} & \begin{tabular}[c]{@{}l@{}}{[}usr{]} I would like to be picked up at the cambridge belfy and go to the cambridge shop.\\ {[}sys{]} I'm sorry, are you going to the Cambridge shop house, \\or did you mean the cambridge shop?\end{tabular}                              & ... \\ \hline
Utt. 1           & \begin{tabular}[c]{@{}l@{}}{[}usr{]} I need to book a taxi to come to Express by Holiday Inn Cambridge \\to take me to the Oak Bistro.\\ {[}sys{]} Okay, waht time do you want to leave by?\end{tabular}     & \begin{tabular}[c]{@{}l@{}}{[}usr{]} I want to be picked up at frankie and bennys please.\\ {[}sys{]} I've booked a black Honda, the contact number is 07796011098.\\ {[}usr{]} Great, thank you for your help.\\ {[}sys{]} No problem. Are you finished?\end{tabular} & ... \\ \hline
Utt. 2           & \begin{tabular}[c]{@{}l@{}}{[}usr{]} I want to depart from sidney sussex college.\\ {[}sys{]} Great. Now I'll just need a time from you please.\end{tabular}                                               & \begin{tabular}[c]{@{}l@{}}{[}usr{]} I am departing from la tasca.\\ {[}sys{]} I have booked a grey BMW, the contact number os 07618837066.\\ {[}usr{]} Thanks so much for your help.\\ {[}sys{]} Thank you goodbye.\end{tabular}                                      & ... \\ \hline
Utt. 3           & ...                                                                                                                                                                                                        & ...                                                                                                                                                                                                                                                                    & ... \\ \hline
\multicolumn{4}{|c|}{\textbf{Restaurant}}\\ \hline
Utt. 0     & \begin{tabular}[c]{@{}l@{}}{[}usr{]} I need help finding a place to eat called curry garden.\\ {[}sys{]} Curry garden is an expensive restaurant that serves Indian food. \\Do you want me to book it for you?\end{tabular} & \begin{tabular}[c]{@{}l@{}}{[}usr{]} Yes, please.I need one for friday at 13:45. I'll be dining alone.\\ {[}sys{]} I apologize but I was unable to book you for the restaurant. \\Would you like for me to find you another restaurant?\end{tabular} & ... \\ \hline
Utt. 1     & \begin{tabular}[c]{@{}l@{}}{[}usr{]} I'm trying to find a restaurant called the Slug and Lettuce. \\Do you know where that is?\\ {[}sys{]} Yes it is in the centre area.\end{tabular}                                          & \begin{tabular}[c]{@{}l@{}}{[}usr{]} 5 people on a Wednesday at 19:45. \\ {[}sys{]} You got it. Here is your reference number: P9D58C0O.\\ {[}usr{]} Thank you for help. That's everything I needed. \\ {[}sys{]} Have a great day!\end{tabular}     & ... \\ \hline
Utt. 2     & \begin{tabular}[c]{@{}l@{}}{[}usr{]} I need to get to a restaurant known as the Lucky star.\\ {[}sys{]} It is located at Cambridge Leisure Park Clifton Way Cherry Hinton.\end{tabular}                                      & \begin{tabular}[c]{@{}l@{}}{[}usr{]} Yes I would, for 3 people Wednesday at 18:15.\\ {[}sys{]} Unfortunately, the restaurant is full at this time. \\Is there another time or day I can reserve for you?\end{tabular}                               & ... \\ \hline
Utt. 3     & ...                                                                                                                                                                                                                          & ...                                                                                                                                                                                                                                                  & ... \\ \hline
\multicolumn{4}{|c|}{\textbf{Hotel}}\\ \hline
Utt. 0     & \begin{tabular}[c]{@{}l@{}}{[}usr{]} I would really like something expensive.\\ {[}sys{]} Unfortunately, I can't find any that are expensive. \\Let's try a different price range.\end{tabular}                                                                                                                                                                                                           & \begin{tabular}[c]{@{}l@{}}{[}usr{]} Yes, please. 6 people 3 nights starting on tuesday.\\ {[}sys{]} I am sorry but I wasn't able to book that for you for Tuesday. \\ Is there another day you would like to stay or perhaps a shorter stay?\end{tabular}                                                                                        & ... \\ \hline
Utt. 1     & \begin{tabular}[c]{@{}l@{}}{[}usr{]} I need a guesthouse with a moderate price. \\ {[}sys{]} Do you have a preferred area of the city you'd like to stay in?\\ {[}usr{]} No, I don't have a preference. \\I'd like the guesthouse to have free parking though!\\ {[}sys{]} I found acorn guest house. It is moderately priced and has four stars! \\ Would you like me to book that for you?\end{tabular} & \begin{tabular}[c]{@{}l@{}}{[}usr{]} Yes. I have heard great things about that guest house. \\ {[}sys{]} Your booking is complete. Your reference number is 33ZFXQ8P. \\ Is there anything else I can help you with today?\\ {[}usr{]} No, that's all I needed today. Thank you!\\ {[}sys{]} Thank you for using our service today."\end{tabular} & ... \\ \hline
Utt. 2     & \begin{tabular}[c]{@{}l@{}}{[}usr{]} I'm looking for something that would be in the middle price point.\\ {[}sys{]} I have located 15 guesthouse types in the moderate price range. \\ Do you have a preference for a particular area of town?\end{tabular}                                                                                                                                             & \begin{tabular}[c]{@{}l@{}}{[}usr{]} That sounds perfect. \\Can I book it for 5 nights starting on Monday for 3 people?\\ {[}sys{]} The Avalon doesn't have room for 5 nights starting Monday. \\How about a different day or a shorter stay?\end{tabular}                                                                                            & ... \\ \hline
Utt. 3     & ...                                                                                                                                                                                                                                                                                                                                                                                                     & ...                                                                                                                                                                                                                                                                                                                                               & ... \\ \hline
\multicolumn{4}{|c|}{\textbf{Attraction}}\\ \hline
Utt. 0     & \begin{tabular}[c]{@{}l@{}}{[}usr{]} Can you please help me find a place to go?\\ {[}sys{]} I've found 79 places for you to go. \\Do you have any specific ideas in mind?\end{tabular}                                                                                                                                                                                                           & \begin{tabular}[c]{@{}l@{}}{[}usr{]} I'd like a sports place in the centre please.\\ {[}sys{]} There are no results matching your query. \\Can I try a different area or type?\end{tabular}                                                                                        & ... \\ \hline
Utt. 1     & \begin{tabular}[c]{@{}l@{}}{[}usr{]} I'm a tourist from out of town. \\But, I was trying to find something fun to do near my hotel. \\Could you recommend a place? \\ {[}sys{]} I would be more than happy to recommend an attraction, \\first could you tell me in what part of town your hotel is located?\end{tabular} & \begin{tabular}[c]{@{}l@{}}{[}usr{]} I'm looking for a place in the centre of town that is a nightclub. \\ {[}sys{]} There are 5 nightclubs in the centre area, \\the one with the lowest entrance fee is soul tree nightclub which is 4 pounds to get in. \end{tabular} & ... \\ \hline
Utt. 2     & \begin{tabular}[c]{@{}l@{}}{[}usr{]} I am looking for a place to go in town.\\ {[}sys{]} There are many places, do you have a particular destination type in mind?\end{tabular}                                                                                                                                             & \begin{tabular}[c]{@{}l@{}}{[}usr{]} Can you found me a swimming pool in the south part of town?\\ {[}sys{]} Unfortunately I was unable to find a pool at the south part of town, \\however we have some north east and at the centre.\end{tabular}                                                                                            & ... \\ \hline
Utt. 3     & ...                                                                                                                                                                                                                                                                                                                                                                                                     & ...                                                                                                                                                                                                                                                                                                                                               & ... \\ \hline

\multicolumn{4}{|c|}{\textbf{Train}}\\ \hline
Utt. 0     & \begin{tabular}[c]{@{}l@{}}{[}usr{]} I am looking for a train departing from london liverpool please.\\ {[}sys{]} I'll be glad to help. \\You would like to from london liverpool street to what destination, please?\end{tabular}                                                                                                                                                                                                           & \begin{tabular}[c]{@{}l@{}}{[}usr{]} I want to leave on Friday.\\ {[}sys{]} Unfortunately, the only train I have matching your criteria is one leaving 23:59 \\and arriving at 01:27 in the morning. \\Did you want to book that?\end{tabular}                                                                                        & ... \\ \hline
Utt. 1     & \begin{tabular}[c]{@{}l@{}}{[}usr{]} I am looking for a train to go to London Kings Cross. \\ {[}sys{]} There are several available option to travel to Kings cross today. \\What time would you like to travel?\end{tabular} & \begin{tabular}[c]{@{}l@{}}{[}usr{]} I want to leave on Monday. \\ {[}sys{]} Great, TR5720 from london kings cross to cambridge leaves monday at 11:17. \\Can I book this for you? \end{tabular} & ... \\ \hline
Utt. 2     & \begin{tabular}[c]{@{}l@{}}{[}usr{]} I am looking for a train from Cambridge to Birmingham New Street.\\ {[}sys{]} The next train leaving from Cambridge for Birmingham New Street \\departs Friday at 5:01, and will arrive by 7:44.\end{tabular}                                                                                                                                             & \begin{tabular}[c]{@{}l@{}}{[}usr{]} I would like to leave after 13:45.\\ {[}sys{]} There is a train that leaves at 15:00 \\would you like me to book that train for you?\end{tabular}                                                                                            & ... \\ \hline
Utt. 3     & ...                                                                                                                                                                                                                                                                                                                                                                                                     & ...                                                                                                                                                                                                                                                                                                                                               & ... \\ \hline
\end{tabular}
}
\caption{Predicted dialogue states for dialogues in the five domains of the MultiWOZ dataset.}
\label{tab:my-table}
\end{table*}
\end{document}